\documentclass[preprint,12pt]{elsarticle}

\usepackage{amssymb}
\usepackage{amsmath}
\usepackage{amssymb}
\usepackage{amsmath}
\usepackage{appendix}
\usepackage{booktabs}
\usepackage{multirow}
\usepackage{adjustbox}
\usepackage{graphicx}
\usepackage{subfig}
\usepackage{url}
\usepackage{xcolor}
\usepackage[figuresright]{rotating}

\journal{International Journal of Approximate Reasoning}

\begin{document}

\begin{frontmatter}

\title{Visual hallucination detection in large vision-language models via evidential conflict\tnoteref{title}}

\author[1,2,3]{Tao Huang\fnref{equal1}}
\author[1,2,3]{Zhekun Liu\fnref{equal1}}
\author[2,4]{Rui Wang\corref{cor1}}
\author[5]{Yang Zhang}
\author[1,2,3]{Liping Jing}

\fntext[equal1]{These two authors contributed equally to this work.}
\cortext[cor1]{Corresponding author. Email: rui.wang@bjtu.edu.cn}
\tnotetext[title]{This paper is a revised and extended version of a paper~\cite{liu2024object} presented at BELIEF 2024.}

\address[1]{\footnotesize Beijing Key Lab of Traffic Data Mining and Embodied Intelligence, Beijing, China}
\address[2]{\footnotesize State Key Laboratory of Advanced Rail Autonomous Operation, Beijing, China}
\address[3]{\footnotesize School of Computer Science and Technology, Beijing Jiaotong University, Beijing, China}
\address[4]{\footnotesize School of Automation and Intelligence, Beijing Jiaotong University, Beijing, China}
\address[5]{\footnotesize School of Electronic and Information Engineering, Beijing Jiaotong University, Beijing, China}

\begin{abstract}
Despite the remarkable multimodal capabilities of Large Vision-Language Models (LVLMs), discrepancies often occur between visual inputs and textual outputs—a phenomenon we term visual hallucination. This critical reliability gap poses substantial risks in safety-critical Artificial Intelligence (AI) applications, necessitating a comprehensive evaluation benchmark and effective detection methods. Firstly, we observe that existing visual-centric hallucination benchmarks mainly assess LVLMs from a perception perspective, overlooking hallucinations arising from advanced reasoning capabilities. We develop the Perception-Reasoning Evaluation Hallucination (PRE-HAL) dataset, which enables the systematic evaluation of both perception and reasoning capabilities of LVLMs across multiple visual semantics, such as instances, scenes, and relations. Comprehensive evaluation with this new benchmark exposed more visual vulnerabilities, particularly in the more challenging task of relation reasoning. To address this issue, we propose, to the best of our knowledge, the first Dempster-Shafer theory (DST)-based visual hallucination detection method for LVLMs through uncertainty estimation. This method aims to efficiently capture the degree of conflict in high-level features at the model inference phase. Specifically, our approach employs simple mass functions to mitigate the computational complexity of evidence combination on power sets. We conduct an extensive evaluation of state-of-the-art LVLMs, LLaVA-v1.5, mPLUG-Owl2 and mPLUG-Owl3, with the new PRE-HAL benchmark. Experimental results indicate that our method outperforms five baseline uncertainty metrics, achieving average AUROC improvements of 4\%, 10\%, and 7\% across three LVLMs. Notably, it exhibits remarkable robustness in scene perception tasks. These results validate that feature-level conflict analysis offers a scalable, cost-effective solution for enhancing LVLM trustworthiness. Our code is available at \url{https://github.com/HT86159/Evidential-Conflict}.
\end{abstract}
\end{frontmatter}


\section{Introduction}
\label{sec:intro}

\begin{figure}[t]
\centering
\includegraphics[page=1,width=\linewidth,clip,trim={0.2cm 0cm 0cm 0cm}]{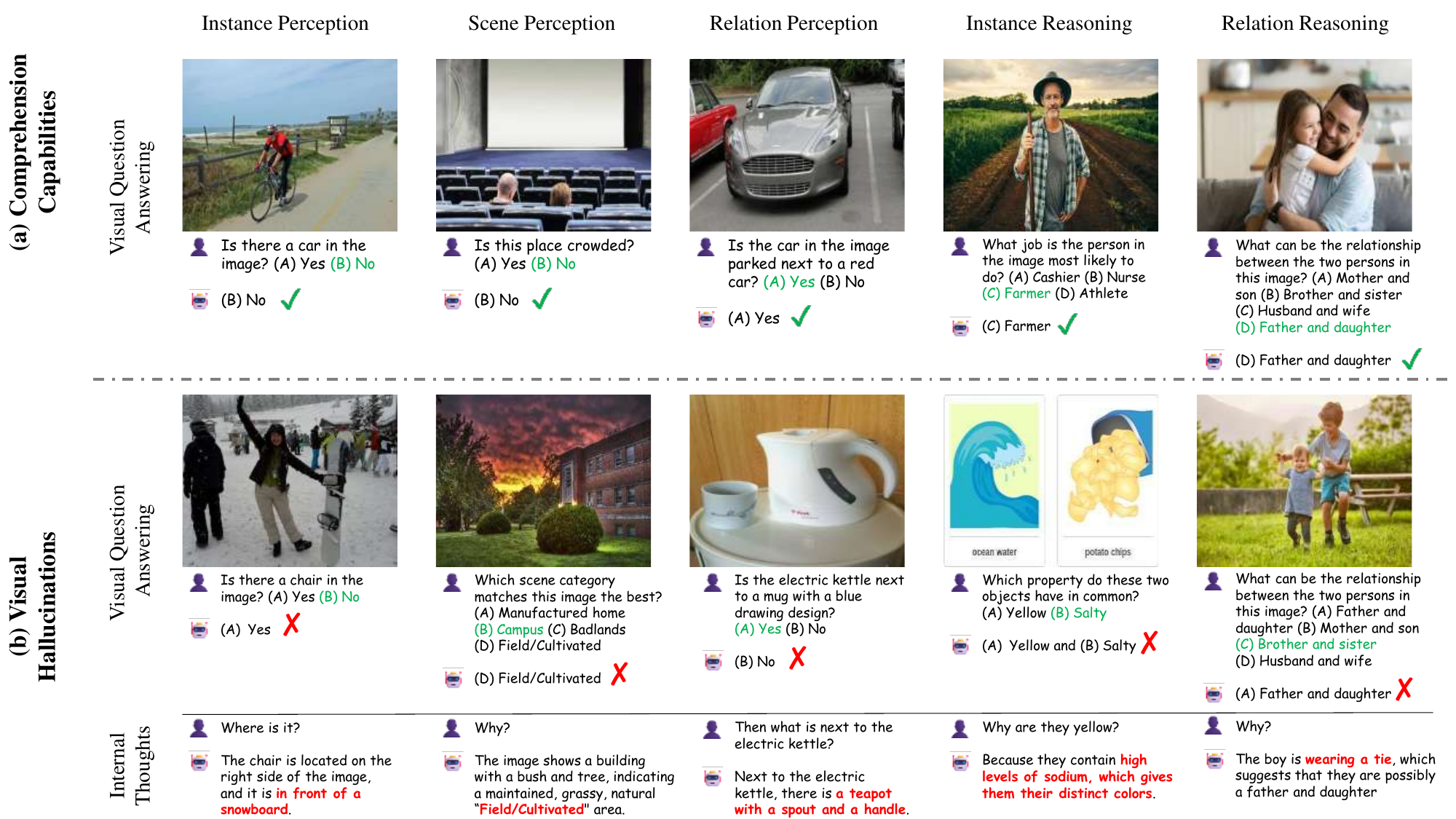}
\caption{
Examples illustrating five types of capabilities of LVLMs (top row) and their corresponding visual hallucinations (bottom row). 
Notably, we further probe the model’s internal thought process behind the generation of visual hallucinations through follow-up questions based on Visual Question Answering.
}\label{fig:hallucination}
\end{figure}

Large Language Models (LLMs) \cite{brown2020language,touvron2023llama,VicunaOpenSourceChatbot,touvron2023llama2} and their multimodal counterparts, Large Vision-Language Models (LVLMs) \cite{daiInstructBLIPGeneralpurposeVisionLanguage2023,liu2024visual,guo2025deepseek}, have achieved near-human-level capabilities in understanding and representing both text and visual data in various tasks, such as text generation \cite{zhang2023survey}, image captioning \cite{lin2014microsoft} and visual question answering (VQA) \cite{antol2015vqa}.  
Despite these advancements, LLMs and LVLMs remain vulnerable to \textit{hallucination} issues\footnote{Researchers also advocate  \textit{confabulation} as an alternative term, emphasizing the generation of erroneous and arbitrary answers~\cite{rawteSurveyTheTroublingEmergence2023,farquhar2024detecting}, due to its potential to promote undue anthropomorphism~\cite{shanahan2024talking}. However, because there is still a lack of consistency in term adoption and the definition of AI hallucinations~\cite{malekiMisnomer2024}, we continue to use \textit{hallucination} in this paper.}~\cite{achiam2023gpt,maynez2020faithfulness,zhang2023siren}, a pervasive challenge intrinsic to generative models.
Particularly, these models struggle with generating false and fabricated information (factuality hallucination) or producing outputs that conflict with input prompts or context (faithfulness hallucination)~\cite{huang2023survey}.
For LVLMs, Huang et al.~\cite{huang2023survey, liu2024survey} refer \textit{visual hallucinations} as discrepancies between the factual content of visual inputs and the generated textual outputs. Although hallucinations are special cases of predictive errors (i.e., mismatches between outputs and ground-truth labels which are commonly caused by unrepresentative data or inductive
biases), they are distinguished by producing outputs that contradict common sense, logical reasoning, or the laws of the physical world. Such hallucinations reflect a model’s misinterpretation of semantic, causal, or physical principles, resulting in nonsensical or physically implausible predictions. Figure~\ref{fig:hallucination} illustrates examples of visual hallucinations across three aspects, including object hallucinations (descriptions of nonexistent objects)~\cite{li2023evaluating}, relation hallucinations (misrepresentation of spatial relationships)~\cite{pmlr-v235-wu24l}, and attribute hallucinations (incorrect interpretations of object attributes)~\cite{hu2023ciem}.

Consequently, these challenges will pose substantial risks in safety- or task-critical applications of these large-scale models, such as intelligent medical diagnosis~\cite{ huangLymphomaSegmentation3D2022,xu2022deep}, autonomous driving~\cite{guo2024divert}, and automated trading~\cite{huang2019automated}.
More alarmingly, the occurrence of hallucinations has been proven to have a statistical lower bound, independent of model architecture or training data~\cite{kalai2024calibrated}. 
This suggests that hallucinations are an inherent characteristic, rather than defects that can be easily rectified.
Thus, until a fundamental solution to this issue is developed, effectively determining whether the outputs of LVLMs are trustworthy and accurately communicate the uncertainty associated with their predictions is crucial for ensuring the reliability of state-of-the-art LVLMs.
Recent studies~\cite{duan2024shifting,cheninside,lingenerating,xiao2021hallucination} reveal a strong correlation between predictive uncertainty and hallucinations in LLMs and LVLMs, driving three types of hallucination detection approaches: verbal elicitation, external consistency checks, and internal uncertainty estimation. Verbal methods~\cite{xiong2024can, tian2023just} prompt models to self-report confidence but often fail due to misinterpretation of instructions or imprecise confidence expression~\cite{kapoor2024large}. 
External methods~\cite{manakul2023selfcheckgpt, raj2023semantic} evaluate output consistency across multiple generations, leveraging external validation models, yet face scalability issues from high computational costs~\cite{kapoor2024large} and dependency on prior knowledge~\cite{farquhar2024detecting,lingenerating}. 
Internal methods traditionally use Monte Carlo dropout~\cite{gal2016dropout} and ensemble method~\cite{lakshminarayanan2017simple}, but these are no longer feasible for large-scale models~\cite{xiao2022uncertainty,kapoor2024large} due to inapplicable dropout layers and excessive resource demands. Alternatives like token-level probability aggregation (negative log-likelihood~\cite{guerreiro2023looking}, entropy~\cite{Malinin2021UncertaintyEI}) or activation analysis based on token importance~\cite{duan2024shifting} are now prioritized. 
Similar to smaller models, both LLMs and LVLMs suffer from overconfidence~\cite{guo2017calibration} and are poorly calibrated, particularly after fine-tuning with reinforcement learning from human feedback (RLHF)~\cite{kadavath2022language,tian2023just}. Consequently, probability-based metrics are limited in detecting erroneous predictions, such as hallucinations.
Therefore, efficient and accurate uncertainty estimation of LVLMs is a key challenge for enhancing their trustworthiness~\cite{xiao2022uncertainty, kapoor2024large, xiong2024can}.

Before presenting our approach for visual hallucination detection, we highlight that existing benchmarks focus primarily on perception, neglecting hallucinations from advanced reasoning. To address this, we introduce the Perception-Reasoning Evaluation Hallucination (PRE-HAL) dataset, which enables systematic evaluation of both perception and reasoning capabilities in LVLMs across various visual semantics, including instances, scenes, and relations.
Evaluation with the new PRE-HAL benchmark exposed more visual vulnerabilities, particularly in the relation reasoning task.
Results show that the corresponding hallucination rates reach 49.44\%,
50.42\%, and 19.92\% for three state-of-the-art LVLMs, LLaVA-v1.5 \cite{liu2024improved}, mPLUG-Owl2 \cite{ye2024mplug2}, and mPLUG-Owl3~\cite{ye2024mplug3}, respectively.

To address the aforementioned issues, we propose, to the best of our knowledge, the first DST-based visual hallucination detection through uncertainty estimation. This approach captures uncertainty in the high-level features of LVLMs at the inference stage. We treat these features as evidence since next-token prediction relies on top-layer representations, typically from feed-forward networks. To reduce computational complexity, we adopt simple mass functions for basic belief assignment, following the approach of~\cite{denoeux2019logistic}. We then combine these mass functions using Dempster's rule to measure evidential uncertainty, which essentially represents feature conflict, avoiding the computational cost of combining evidence across power sets.
We evaluate our approach thoroughly in the VQA tasks, and experimental results show that our method outperforms the best intern baseline methods for visual hallucination detection, achieving an AUROC improvement of 4\%, 10\% and 7\% on LLaVA-v1.5 \cite{liu2024improved}, mPLUG-Owl2 \cite{ye2024mplug2}, and mPLUG-Owl3~\cite{ye2024mplug3} particularly under conditions of high uncertainty. Moreover, the proposed method shows consistently robust performance across the five fine-grained hallucination types in the new PRE-HAL dataset, particularly excelling in the scene perception type.

\section{Related Work}
In this section, we present key concepts integral to the comprehension of our work. Specifically, in Subsection \ref{sec:ds}, a brief background on Dempster-Shafer Theory (DST) is presented. Simultaneously, Subsection \ref{sec:lvlms} provides an overview of the hallucination detection methods applicable to Large Vision-Language Models (LVLMs).
\subsection{Dempster Shafer Theory}\label{sec:ds}
The Dempster-Shafer Theory (DST), proposed by Dempster \cite{dempster1967upper} and Shafer \cite{shafer1976mathematical}, extends classical probability theory to manage uncertainty and partial belief. Unlike traditional probability, which assigns definite probabilities to events, DST employs basic probability assignments (BPA). It distributes belief among subsets of the sample space, allowing for a fine-grained representation of uncertainty. 
This theory can fuse evidence from different sources using Dempster's rule of combination. 
Now, we recall the key concepts and definitions applied in this paper.
\subsubsection{Fundamentals of DST}
Let \( \mathcal V = \{z_1, z_2, \dots, z_I \} \) represent the frame of discernment, which contains $I$ possible outcomes. In this context, a \textit{mass function} \( m(\cdot) \) maps subsets of the frame of discernment \( 2^{\mathcal V} \) to the interval \([0, 1]\), indicating the degree of belief assigned to each subset. The mass function is subject to the normalization condition,
\begin{equation}
    \sum_{A \subseteq \mathcal V} m(A) = 1; \quad m(\emptyset)=0,
\end{equation}
where \( A \) is any subset of \( \mathcal{V} \), and \( \emptyset \) represents the empty set.
For a subset \( A \subseteq\mathcal V \), if \( m(A) > 0 \), \( A \) is called a \textit{focal set} of \( m(\cdot) \). 
Specifically, a mass function is called \textit{simple} when it assigns belief exclusively to a specific subset \( A \subseteq \mathcal{V} \) and the \( \mathcal{V} \). Formally, it is defined as follows:
\begin{equation}
 m(A)=s;\quad m(\mathcal V)=1-s,
\end{equation}
where \( A \neq \emptyset \) and $s \in [0,1]$ represents the \textit{degree of support} for A.
In particular, the mass \( m(\mathcal V) \), assigned to the entire frame, commonly indicates the \textit{degree of ignorance}, as it exhibits no preferential allocation towards any particular subset. 
Given two mass functions, $m_{1}(\cdot)$ and $m_{2}(\cdot)$, which represent evidence from two different sources (e.g., agents), the combined mass function for all $A \subseteq \mathcal V$, with $A \neq \emptyset$, is computed through Dempster's rule~\cite{shafer1976mathematical} as follows:
\begin{equation} 
 \left(m_1 \oplus m_2\right)(A)=\frac{1}{1-\kappa} \sum_{B \cap C=A} m_1(B) m_2(C);\quad \kappa=\sum_{B \cap C=\emptyset} m_1(B) m_2(C),
\end{equation} 
where $\left(m_1 \oplus m_2\right)(\emptyset)=0$, and the $\kappa$ serves as an important metric to measure the \textit{degree of conflict} between \( m_1(\cdot) \) and \( m_2(\cdot) \).

Given a mass \( m(\cdot) \), two useful functions, the \textit{belief} and \textit{plausibility functions}, are defined, respectively, as
\begin{equation} 
Bel(A)=\sum_{B\subseteq A} m(B);\quad Pl(A)=\sum_{B\cap A\not= \emptyset}m(B).
\end{equation}
The \textit{belief function} \( Bel(A) \) represents the degree of certainty that the true state lies within the subset \( A \) based on all available evidence, excluding any possibility outside of \( A \). 
In contrast, the \textit{plausibility function} \( Pl(A) \) indicates the degree of belief that the true state may lie within \( A \), without ruling out possibilities.

In the case of singletons (i.e., \( \{z_1\} \)), the plausibility function \( Pl(\cdot) \) is restricted to the \textit{contour function} \( pl(\cdot) \) (i.e., \( pl(z_i) = Pl(\{z_i\}), \;\forall z_i \in \mathcal{V}\)). The contour function \( pl(z_i) \) measures the plausibility of each singleton hypothesis and assesses the  uncertainty of each possible outcome independently. Furthermore, given two contour functions \( pl_1(\cdot) \) and \( pl_2(\cdot) \), associated with mass functions \( m_1(\cdot) \) and \( m_2(\cdot) \) respectively, they can be combined as
\begin{equation} \label{equ:plcombine}
pl_1\oplus pl_2(z_i)=\frac{1}{1-\kappa} pl_1(z_i)pl_2(z_i),\quad \forall z_i \in \mathcal V.
\end{equation} 
This combination rule simplifies evidence aggregation by directly multiplying the plausibilities of singletons, making the process more efficient.

\subsubsection{Evidential Neural Networks}
DST offers significant advantages in modeling and fusing uncertain information~\cite{denoeux2019decision,deng2015generalized}.
A pioneering study of DST for capturing prediction uncertainty in neural networks proposed by Den\oe ux~\cite{denoeux2000neural} measure the distances to prototypes of an input as evidence and adopts a discounted Bayesian mass function with focal sets limited to only singletons and the frame of discernment.
Later, the same researcher~\cite{denoeux2019logistic} views the high-level features from neural network classifiers as evidence and proposes to use the simple mass function to model this evidential information. 
This approach provides a new interpretation of the output class probabilities from the softmax layer in a neural network classifier.
Specifically, the probabilities are viewed as normalized plausibilities derived from the
simple mass functions, which enabling a principled characterization of model prediction uncertainty.

Owing to the nice properties of the simple mass function for evidence combination, this work alleviate the high computational cost associated with the evidence combination using Dempster’s rule.
Notably, the complexity of evidence fusion grows exponentially
with the size of the frame of discernment, which limits its practical use in real-world applications~\cite{HAENNI2002103,VOORBRAAK1989525}.
As a result, it opens up possibilities for uncertainty-aware learning and inference in large-scale neural networks, with applications in medical diagnosis~\cite{huangLymphomaSegmentation3D2022}, driving scenario understanding~\cite{wang2025reliable}, and hallucination detection for LVLMs, as explored in this paper.


\subsection{Large Vision-Language Models and Hallucination} 
\label{sec:lvlms}
Large Vision-Language Models (LVLMs) integrate the capabilities of large language models (LLMs) and vision models, achieving near-human-level performance in understanding and representing both textual and visual data across diverse tasks. 
Despite their success, LVLMs still face serious challenges, including hallucination~\cite{rohrbach2018object}. The following sections will introduce LVLMs and discuss these hallucination challenges.

\subsubsection{Large Vision-Language Models}
Motivated by the progress of LLMs~\cite{blum2024learning,chang2024survey}, researchers have developed powerful Large Vision-Language Models (LVLMs) by integrating visual encoders with LLMs. 
The remarkable capabilities of LVLMs emerge from three key factors: (1) the massive scale of parameters and training data~\cite{kaplan2020scaling}; (2) architectural innovations, such as the attention mechanisms~\cite{vaswani2017attention}; and (3) advancements in self-supervised pre-training using unlabeled text corpora \cite{devlin2019bert}.

Specifically, as shown in Figure \ref{fig:framework}a, LVLM architectures typically consist of four functional components: a visual encoder, a text encoder, a modality fusion module, and a language model decoder \cite{duSurveyVisionLanguagePreTrained2022}.
The visual encoder, commonly based on the vision encoders from CLIP~\cite{radfordLearningTransferableVisual2021} or DINOv2~\cite{oquabdinov2}, transforms input images into visual tokens.
The modality fusion module bridges vision and language by aligning visual tokens with the semantic space of LLMs.
There are various forms for modality fusion module, including cross-attention~\cite{alayrac2022flamingo}, adapters~\cite{gao2023llama}, simple linear transformations or multilayer perceptron (MLP)~\cite{liu2024visual,liu2024improved}. 
The text encoder and language model decoder serve as the core processing components in LVLMs. These modules receive aligned visual and textual representations and integrate them to generate coherent responses. Well-known LLMs include LLaMA~\cite{touvron2023llama} and Vicuna~\cite{VicunaOpenSourceChatbot}.

Despite their success, LVLMs still confront several challenges such as hallucinations~\cite{rohrbach2018object,biten2022let,li2023evaluating}, which will be discussed in the following subsection.  
\subsubsection{Hallucination of Large Vision-Language Models}
\label{sec:hallucination}
The architectural integration of LLMs into LVLMs leads to inherited hallucination patterns, as discussed by ~\cite{li2023evaluating}.
However, the introduction of visual modalities and multimodal tasks~\cite{karpathy2015deep,antol2015vqa} has rendered traditional LLM-based hallucination characterizations insufficient for LVLMs.

Current research has identified three prominent types of hallucinations in LVLMs: (1) object hallucination, where models describe non-existent objects in images~\cite{liu2024object}, (2) relation hallucination, where models fail to recognize inter-object relationships~\cite{pmlr-v235-wu24l} accurately, and (3)  attribute hallucination~\cite{hu2023ciem}, where models assign incorrect attributes to objects in the image.
To thoroughly characterize hallucinations in LVLMs, we refer to visual hallucination as the discrepancies between the factual content of visual information and the corresponding generated textual content. This encompasses untruthful information~\cite{liu2024object,pmlr-v235-wu24l,hu2023ciem}, as illustrated in Figure \ref{fig:hallucination}.

Such hallucinations may significantly degrade model performance, severely harming user experience in real-world applications~\cite{ji2023survey}, and even posing serious risks to safety-critical applications such as intelligent transportation~\cite{guo2024divert}, and medical imaging analysis~\cite{ huangLymphomaSegmentation3D2022,xu2022deep}.
Therefore, comprehensive evaluation of LVLMs and effective hallucination detection are key solutions to ensuring the accuracy and consistency of outputs.

\subsection{Hallucination Detection Methods}
Due to the large number of parameters and the high training cost of LVLMs, traditional uncertainty quantification methods, such as Monte Carlo Dropout~\cite{gal2016dropout} and ensemble methods~\cite{lakshminarayanan2017simple}, are not suitable for these models. 
As a result, the field of hallucination detection in LVLMs remains in its early stages. Currently, the main approaches rely on classical methods, such as probability or entropy.

\subsubsection{Verbal Elicitation}
Zhang et al. \cite{zhang2024r} argue that when a question falls outside the scope of the model's parametric knowledge, it is more likely to generate hallucinations. 
Therefore, it is crucial for models to express the probability of their outputs.
Verbal elicitation methods typically generate complex outputs that include both answers and their associated probabilities.  
A common approach is to directly use prompting strategies \cite{xiong2024can,tian2023just} to elicit verbalized probabilities, for example, based solely on the simple prompt: \texttt{Provide the probability that your answer is correct. Give ONLY the probability, no other words or explanation.}
Additionally, \cite{linteaching} empirically shows that the uncertainty expressed in words and that extracted from model logits are broadly similar.
However, the model may not always follow the prompt instructions, as failures can occur during its internal handling of the prompt structure, making it challenging to construct partial results~\cite{kapoor2024large}. 
\subsubsection{Internal Method}
Internal methods are often based on the internal information of the model to evaluate confabulation problems. Such methods rely on the probability distribution of the softmax layer to quantify uncertainty with the help of metrics such as negative log probability~\cite{guerreiro2023looking} and entropy~\cite{Malinin2021UncertaintyEI}. 
Furthermore, Kadavath et al.~\cite{kadavath2022language} take the probability a model assigns to the proposition that a specific sample is the correct answer to a question as an uncertainty measure.
Additionally, \cite{duan2024shifting} proposes the Shifting Attention to More Relevant (SAR) components at both the token and sentence levels to improve uncertainty by weighting each token according to its importance.
However, the softmax function suffers from the overconfidence problem~\cite{gal2016dropout,guo2017calibration}, so these uncertainty measures often fail to accurately reflect the true confidence of the model, which may lead to limited accuracy in hallucination detection.
\subsubsection{External Method}
In many practical application scenarios, users often cannot access the internal information of the model (such as in API calls), so external methods are more practical. Farquhar et al.~\cite{farquhar2024detecting} proposed to evaluate the semantic similarity between multiple outputs generated by the model by introducing semantic entropy. Raj et al.~\cite{raj2023semantic} and Manakul et al.~\cite{manakul2023selfcheckgpt} proposed that the semantic consistency of multiple responses to the same question by the analytical model can also be used for hallucination detection. 
Furthermore, Lin et al.~\cite{lingenerating} use pairwise similarity scores to estimate uncertainty.
These methods identify hallucinations by evaluating the semantic consistency or similarity of outputs without accessing the internal information of the model, making them more generalizable in practical applications.

\section{Perception-Reasoning Hallucination Evaluation Dataset}\label{sec:benchmark}
In this section, we will provide a detailed introduction to the proposed visual hallucination evaluation benchmark, Perception-Reasoning Hallucination Evaluation Dataset (PRE-HAL) for LVLMs. The dataset is publicly available at \url{https://huggingface.co/datasets/thuang5288/PRE-HAL}.
\subsection{Lack of Comprehensive Visual Hallucination Benchmark}
Existing LVLM hallucination benchmarks are widely adopted to evaluate LVLMs from the perspective of visual semantics \cite{liu2024survey}, which involves objects, scenes, and relationships, such as the Polling-based Object Probing Evaluation (POPE)~\cite{li2023evaluating}, Negative Object Presence Evaluation (NOPE)~\cite{lovenia-etal-2024-negative}, and Relationship Hallucination Benchmark (R-Bench)~\cite{pmlr-v235-wu24l}. 
However, these benchmarks primarily focus on a single aspect of the model's perceptual capabilities, overlooking the potential for hallucinations during the inference phase. 
As shown in Figure \ref{fig:hallucination}, LVLMs exhibit different kinds of capabilities (the upper row), which can directly lead to specific types of hallucinations (the lower row).
Recently, Liu et al. proposed MMBench \cite{liu2024mmbench}, which emphasizes perception and reasoning as core abilities and further subdivides them into 20 finer dimensions, including attribute recognition, spatial relationships, etc.
In addition, \cite{tong2024eyes} introduced the Multimodal Visual Patterns (MMVP) benchmark, which includes images that CLIP misinterprets, exposing the limitations of LVLMs in handling visually distinct patterns.
Moreover, the variation in data formats and evaluation metrics across datasets complicates an extensive assessment of hallucinations in LVLMs \cite{liu2024mmbench}.
To fill this gap, we propose a comprehensive benchmark for LVLMs' hallucination detection, PRE-HAL.

\subsection{Dataset Construction}

\begin{figure}[t!]
\centering
\includegraphics[page=1,width=0.8\linewidth,clip, trim={0cm 4.5cm 0cm 5cm}]{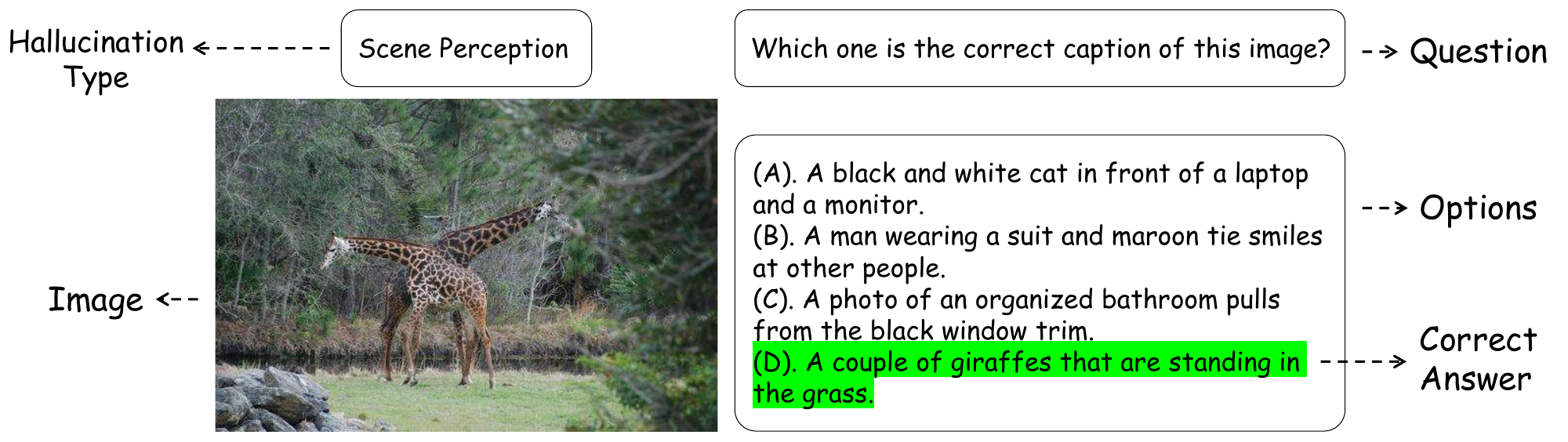}
\caption{The components of a data entry.}\label{fig:example}
\end{figure}

\begin{figure}[htbp]
    \hspace{1cm} 
    \centering
    \subfloat[Visual hallucination data sources for each capability dimension.]{%
        \begin{minipage}[b]{0.5\textwidth}
            \centering
            
                \includegraphics[width=\textwidth,clip, trim={6cm 20cm 2.5cm 2.7cm}]{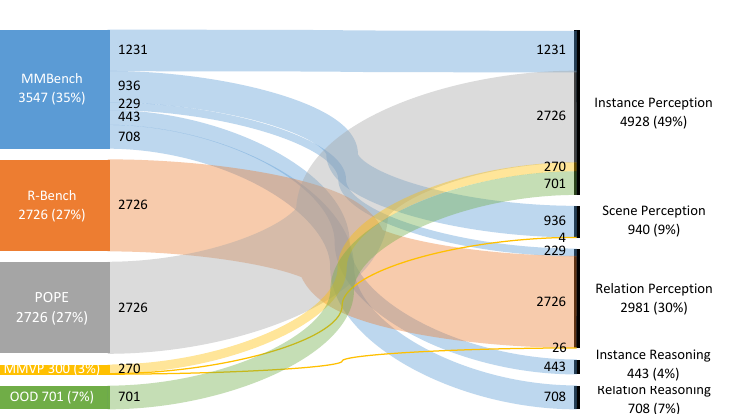}%
            
        \end{minipage}%
        \label{fig:sankey}%
    }
    \hspace{0cm} 
    \subfloat[Capability dimensions in PRE-HAL.]{%
        \begin{minipage}[b]{0.38\textwidth}
            \centering
            \includegraphics[width=\textwidth,clip, trim={3.2cm 2.3cm 3.2cm 2.3cm}]{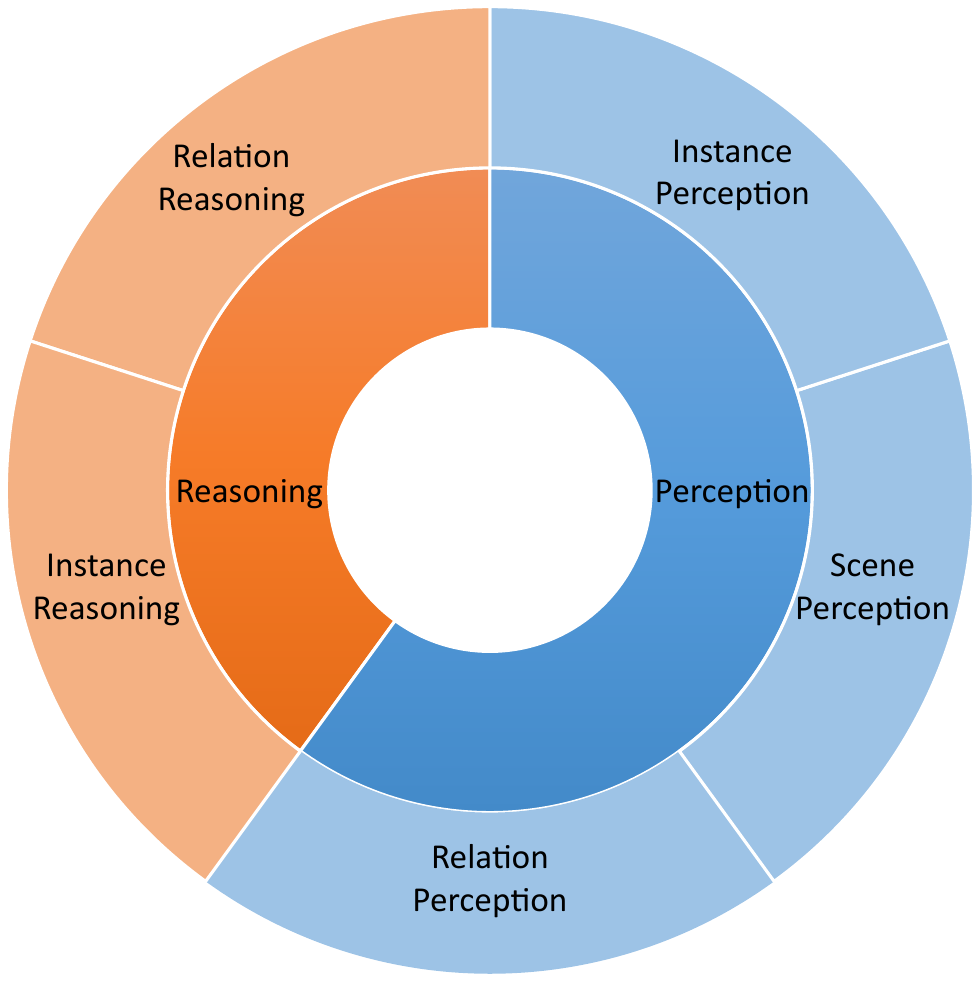}
        \end{minipage}%
        \label{fig:xuri}%
    }
    \caption{The composition and the hierarchical capability dimensions of PRE-HAL.}
    \label{fig:subfloat-combined}
\end{figure}

\paragraph{Data Composition}
Similar to other LVLM benchmarks, PRE-HAL collects VQA data in the form of multiple-choice questions for efficient statistics and evaluation, enabling inference tests on LVLMs.
Inspired by the categorization of multimodal understanding capabilities of LVLMs in MMBench \cite{liu2024mmbench}, we define our benchmark dimensions around perception and reasoning, where perception involves gathering sensory information, and reasoning draws conclusions from that information. We also incorporate visual semantics \cite{liu2024survey} by considering instances, scenes, and relations.
As shown in Figure~\ref{fig:example}, each data entry in PRE-HAL comprises five components: an image, a question, 2 to 4 options, the correct answer, and the hallucination type.
This categorization provides a systematic approach for detecting visual hallucinations in LVLMs, aiding in more precise performance assessment.
It is important to note that the question should be visually dependent, rather than a visual supplement~\cite{guan2024hallusionbench}. Visual supplement questions are those that can be answered without relying on visual input (e.g., \texttt{Which country's capital is Paris?}), where the visual input serves only as complementary information to the text.
Finally, all data entries have been verified by human evaluators.
\paragraph{Data Collection}
We collected multiple data entries for each type of hallucination from a variety of data sources \cite{liu2024mmbench,tong2024eyes,li2023evaluating,pmlr-v235-wu24l}.
The data format in MMBench and MMVP aligns with our expectations, making it directly applicable.
Since POPE and R-Bench are designed to induce object and relationship hallucinations, respectively, with answers limited to \texttt{yes} or \texttt{no}, we converted them into a multiple-choice question format.
Out-of-Distribution (OOD) data, which lies outside the training data distribution of LVLMs, is more likely to induce hallucinations \cite{huang2021therapeutics}. 
To enhance the completeness of PRE-HAL, we manually constructed a set of OOD data and incorporated it into our benchmark.
Specifically, we selected nouns representing objects rarely encountered in daily life, such as biomedical terms from the Therapeutics Data Commons (TDC)~\cite{huang2021therapeutics}, and combined common nouns in pairs to create uncommon word compounds, like ``sand monkey," to generate the questions.
Additionally, the MCOCO image dataset~\cite{LinMScoco2014} was selected to create the image-text pairs.
Different data sources contain various types of hallucinations. The composition sources for each type of visual hallucination are illustrated in Figure~\ref{fig:sankey}.
Last but not least, we manually annotated the hallucination types for all data.

\subsection{Comprehensive Visual Hallucination Evaluation}
\begin{figure}
\centering
\includegraphics[page=1,width=1\linewidth]{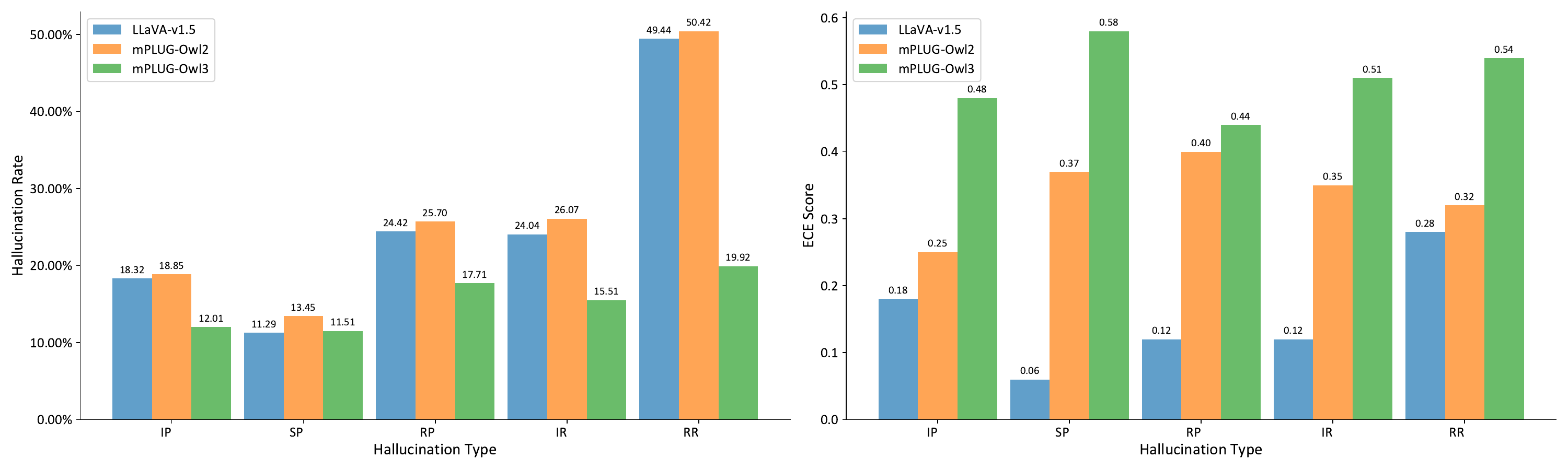}
\caption{Hallucination rates and ECE scores on five types of hallucinations.}\label{fig:rates and scores}
\end{figure}

We evaluated three advanced LVLMs,  LLaVA-v1.5~\cite{liu2024visual}, mPLUG-OWL2~\cite{ye2024mplug2} and mPLUG-OWL3~\cite{ye2024mplug3}, using PRE-HAL.
As shown in the first example of Figure~\ref{fig:hallucination}b, the LVLM incorrectly identifies an instance in the image as a `chair', leading to the selection of the wrong answer. In contrast, as shown in the last example of Figure~\ref{fig:hallucination}b, due to the mis-identification of the tie of one individual in the image, model erroneously infers that the boy is the father of the baby girl.
Statistically, the results show that the hallucination rate for LLaVA-v1.5 is $23.66\%$, for mPLUG-OWL2 is $22.71\%$, and for mPLUG-Owl3, the hallucination rate is $14.38\%$.
Specifically, the incidence of each hallucination type is presented on the left side
of Figure~\ref{fig:rates and scores}. Furthermore, to assess the overconfidence levels of LVLMs, we calculated their Expected Calibration Error (ECE) scores~\cite{guo2017calibration}. The ECE scores for LLaVA-v1.5, mPLUG-Owl2, and mPLUG-Owl3 are 0.35 (blue), 0.19 (orange), and 0.48 (green), respectively. The elevated ECE score of mPLUG-Owl3 indicates that this model alleviates the hallucination issue at the cost of increased overconfidence, thereby introducing potential risks to high-confidence prediction errors.
The ECE scores for each type of hallucination are shown on the right side of Figure~\ref{fig:rates and scores}.

Through thorough evaluations, we identified that reasoning-based tasks, such as those requiring logical inference or complex decision-making, pose greater challenges for LVLMs than perception-based tasks. As a result, the incidence of reasoning-related hallucinations is significantly higher than perception-based ones in both models. Furthermore, the occurrences of scene, instance, and relation hallucinations increase progressively. These findings not only reveal the non-negligible hallucination issues in current poorly calibrated LVLMs but also highlight the urgent need for an effective and scalable hallucination detection method to ensure the trustworthiness of state-of-the-art LVLMs.

\section{Visual Hallucination Detection via Evidential Conflict}
\label{sec:Method}

\begin{figure}
\centering
\includegraphics[width=1\linewidth,clip, trim={0cm 0cm 0cm 0cm}]{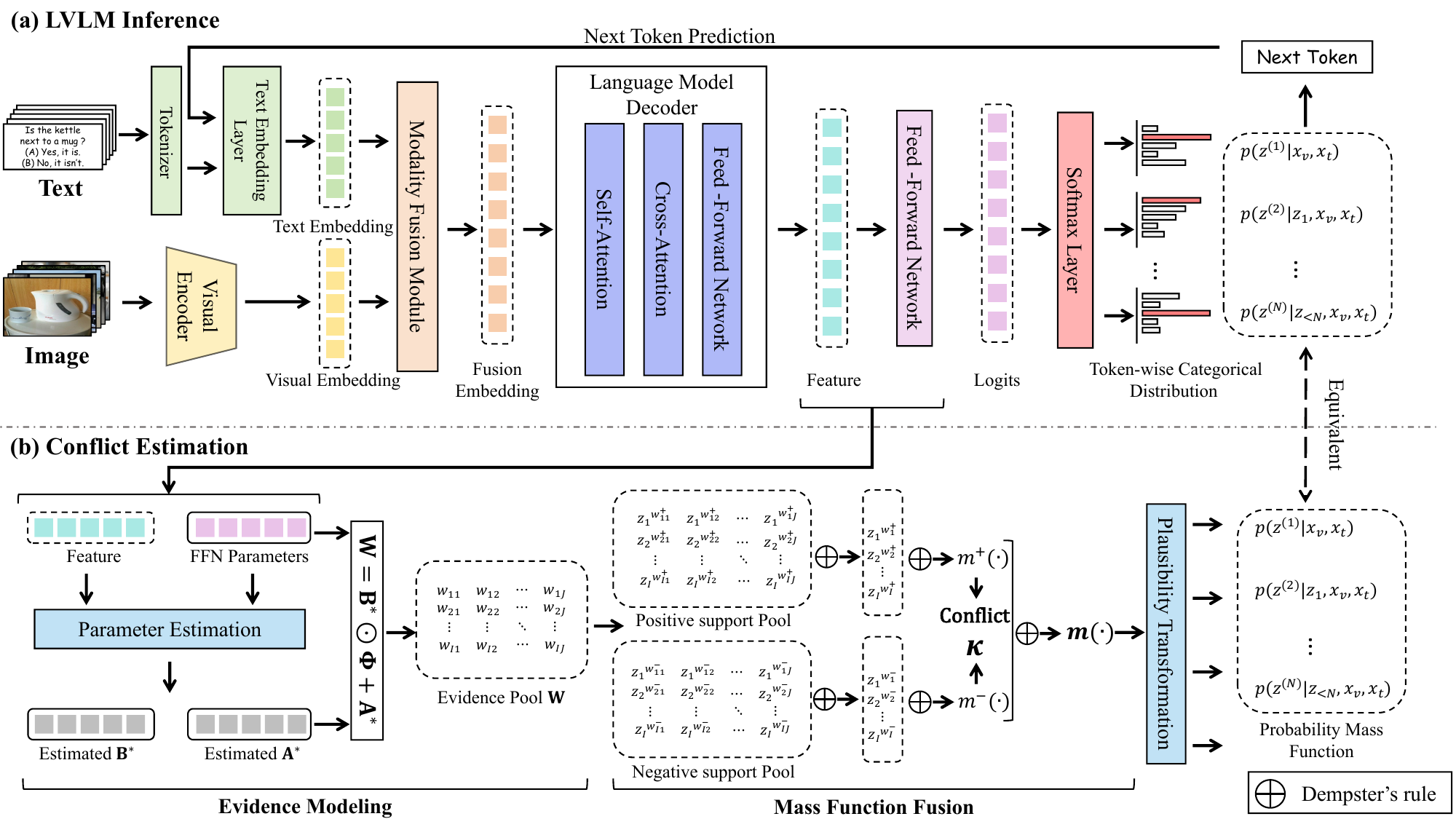}
\caption{Scheme of the proposed method. The upper subfigure illustrates the next-token inference process of LVLMs. The lower subfigure shows the conflict estimation process, where FFN parameters and features serve as inputs, and the conflict $\kappa$ is the output.}
\label{fig:framework}
\end{figure}
In visual-language tasks such as VQA, uncertainties that lead to visual hallucinations in LVLMs can arise from both intra-modality and inter-modality sources. Intra-modality uncertainty occurs within a single modality (e.g., low-resolution images or vague textual descriptions), while inter-modality uncertainty stems from misalignment between the two modalities (e.g., when they provide contradictory or inconsistent information). These uncertainties, whether due to a lack of supporting information or contradictory data, ultimately manifest as discrepancies in the high-level fused features from both modalities. As a result, capturing these conflicts can enhance the effectiveness of the detection of visual hallucinations. Thus, in this section, we present a novel visual hallucination detection based on evidential conflict.
First, we introduce the token prediction process in Subsection \ref{sec:next_token}, followed by a discussion of evidence modeling in Subsection \ref{sec:evidencemodeling}, and the evidence combination and conflict estimation in Subsection \ref{sec:massfusion}.
\subsection{Next-Token Prediction within Large Vision-Language Models}\label{sec:next_token}
As depicted in Figure~\ref{fig:framework}a, the text and image inputs are first processed by the text embedding layer and visual encoder, respectively, generating separate embeddings for each modality. These embeddings are then passed into the modality fusion module, which combines them into a unified fusion embedding. This fused representation is fed into the LLM decoder, where it generates features that are then processed through the Feed-Forward Network (FFN) and softmax layers to predict the next token. The output yields a token-wise categorical distribution, from which the token with the highest probability is selected as the next token. This token is then concatenated with the original text and passed back through the text embedding layer for the next round of inference, continuing until a stop condition is met (e.g., when an end-of-sequence token is generated or the maximum token length is reached).

Typically, LVLMs with a finite token vocabulary \( \mathcal{V}=\{z_1,z_2,\dots,z_I\} \) (e.g., words in English) are designed to predict the $n$-th token \( z^{(n)} \) based on the preceding tokens \( z^{(<n)} \), visual input \( x_v \), and textual input \( x_t \), as described by
\begin{equation}
\begin{aligned}
    z^{(n)}=\underset{z^{(n)}\in \mathcal V}{\arg\max}\;p(z^{(n)}|z^{(<n)},x_v,x_t).
\end{aligned}
\end{equation}

Let \( \mathbf{\Phi} = (\phi_1, \phi_2, \dots, \phi_J)^\top \) represent the high-level feature vector with dimensionality \( J \times 1 \). The parameter matrix of the penultimate layer (i.e., the Feed-Forward Network, FFN) is denoted as \( \widehat{\mathbf{B}} \), and the bias vector is denoted as \( \widehat{\mathbf{A}} \). The final output probability distribution for the next token generated by an LVLM can be expressed as

\begin{equation}
\begin{aligned}\label{equ:lvlmssoftmax}
    \mathbf{p}(z|z^{(<n)},x_v,x_t)=\frac{\exp(\widehat{\mathbf{B}}\mathbf{\Phi}+\widehat{\mathbf{A}})}{\Vert \exp(\widehat{\mathbf{B}}\mathbf{\Phi}+\widehat{\mathbf{A}})\Vert_1};\quad n =1,2,\dots,N,
\end{aligned}
\end{equation}
where \( \Vert \cdot \Vert_1 \) denotes the \( L_1 \)-norm, which is the sum of the absolute values of the components of the vector.

\subsection{Evidence Modeling}\label{sec:evidencemodeling}
High-level features are crucial for model decisions, such as next-token prediction in LVLMs. As discussed previously, effectively capturing conflicts in these features enhances, to a great extent, the performance of visual hallucinations detection. Therefore, we treat the high-level features \( \mathbf{\Phi} \) as evidence, using them to perform basic belief assignment, and ultimately measure the conflict of this evidence. Following the interpretation in \cite{denoeux2019logistic} of the softmax layer calculations within an FFN as a combination of simple mass functions based on \( \mathbf{\Phi} \), we consider the conflict within \( \mathbf{\Phi} \) as an uncertainty metric for visual hallucination detection. This conflict is measured using the process outlined in Figure~\ref{fig:framework}b.

Building on this framework, the support provided by each feature \( \phi_{j} \) is assigned to either the singleton \( \{z_{i}\} \) or its complement \( \overline{\{ z_{i} \}} \), depending on the sign of weight of evidence $w_{ij}$ associated with simple mass function $m_{ij}(\cdot)$, this can be represented by
\begin{equation}\label{equ:linearweights}
\mathbf{W}=\mathbf{B}\odot\mathbf{\Phi}^\top+\mathbf{A},
\end{equation}
where \( \mathbf{W} \) denotes the evidence pool matrix, with elements \( \{w_{ij}\}_{i=1,j=1}^{I,J} \). Additionally, \( \mathbf{B} \) and \( \mathbf{A} \) are parametric matrices to be estimated, with elements \( \{\beta_{ij}\}_{i=1,j=1}^{I,J} \) and \( \{\alpha_{ij}\}_{i=1,j=1}^{I,J} \), respectively.
Specifically, the positive and negative components of \( \mathbf{W} \), denoted as \( w_{ij}^{+} = \max(0, w_{ij}) \) and \( w_{ij}^{-} = \max(0, -w_{ij}) \), are assigned to \( \{z_{i}\} \) and its complement \( \overline{\{ z_{i} \}} \). Thus, the corresponding simple mass functions are
\begin{equation} \label{eq:m_ij}
    m_{ij}^{+}(\cdot):=\left\{ z_{i} \right\}^{w_{ij}^{+}};\quad m_{ij}^{-}(\cdot):=\overline{ \left\{ z_{i} \right\} }^{w_{ij}^{-}},\quad i=1,2,\dots,I;~j=1,2,\dots,J,
\end{equation}
representing the degree of support from \( \phi_j \) to \( z_i \), and \( \overline{\{ z_i \}} \), respectively.


Next, the parameters \( \mathbf{B} \) and \( \mathbf{A} \) are estimated according to the Least Commitment Principle (LCP) \cite{smets1993belief}. 
Specifically, when multiple mass functions are compatible with a given set of constraints, the goal is to avoid making unnecessary assumptions or overcommitting. 
This is achieved by determining the parameters \( \mathbf{B} \) and \( \mathbf{A} \) that yield output mass functions with minimal informational content. The optimization problem can be formally expressed as
\begin{equation}\label{equ:lcp}
\begin{aligned}
&\min_{\mathbf{B},\mathbf{A}} \quad\Vert\mathbf{B}\odot\mathbf{\Phi}^\top+\mathbf{A}\Vert_2^2,\\
\text{s.t.}\; & \mathbf{1}_{1\times I}\mathbf{A} =\mathbf{\widehat{{\beta}}_{0}}+c_0;\quad \mathbf{B}=\widehat{\mathbf{B}}+\mathbf{c},
\end{aligned}
\end{equation}
where \( \mathbf{1}_{1 \times I} \) denotes the all-ones vector of dimension \( 1 \times I \),  \( \mathbf{\widehat{{\beta}}_{0}}\) is the parameter from the FFN, and \( c_0 \) and \( \mathbf{c} \) are auxiliary parameters to be estimated. 
In \cite{denoeux2019logistic}, the closed-form solutions for the auxiliary parameters, \( \mathbf{B} \), and \( \mathbf{A} \) were
shown to have the following expression:
\begin{equation}\label{equ:bstart}
c_0=-\frac1J \mathbf{\widehat{{\beta}}_{0}}\mathbf{1}_{J\times J};\quad \mathbf{c}=-\frac{1}{I} \mathbf{1}_{1 \times I}\widehat{\mathbf{B}} ;\quad\mathbf{B}^* = \underbrace{\widehat{\mathbf{B}} - \frac{1}{I} \mathbf{1}_{1 \times I}\widehat{\mathbf{B}}}_{\text{Row centering}};
\end{equation}
\begin{equation}\label{equ:astart}
\mathbf{A}^*=\frac1I(\mathbf{1}_{I\times 1}\mathbf{\beta_0^*})-(\underbrace{\mathbf{B}^*-\frac1J(\mathbf{B}^*)^\top\mathbf{1}_{J\times J}}_{\text{Column centering}})\odot\Phi^\top
\end{equation}
where $\mathbf{\beta_0^*}=\mathbf{\widehat{{\beta}}_{0}}-\frac1J \mathbf{\widehat{{\beta}}_{0}}\mathbf{1}_{J\times J}$ is a $1\times J$ vector.
It is important to note that \eqref{equ:bstart} essentially performs row-wise centering on the original model weights, where the mean of each row is subtracted. This operation preserves the relative magnitudes of the logits while reducing the overall commitment.
Additionally, \eqref{equ:astart} averages the original model bias and subtracts the inner product of the double-centered model weights and the features. This operation also contributes to the reduction of commitment.

\subsection{Mass Function Fusion and Conflict Estimation}\label{sec:massfusion}
To obtain the overall conflict degree \( \kappa \) within the feature $\mathbf{\Phi}$, we need to combine all simple mass functions associated with $\mathbf{\Phi}$.
Since we have the evidence pool \( \mathbf{W} \), which contains all the weights of evidence, the corresponding simple mass function can be expressed as~\eqref{eq:m_ij}.
We denote the simple mass function with focal set $A$ and weight of evidence $w$ as $A^w$. Consider two simple mass functions $A^{w_1}$ and $A^{w_2}$, which have degrees of support $s_{1}$ and $s_{2}$ respectively. The combined mass function is then given by
\begin{equation} 
\left(A^{w_1} \oplus A^{w_2}\right)(A)  =1-\left(1-s_1\right)\left(1-s_2\right);\quad
\left(A^{w_1} \oplus A^{w_2}\right)(\mathcal V)  =\left(1-s_1\right)\left(1-s_2\right).
\end{equation} 
The combined weight of evidence $w_{12}$ associated to $A^{w_1} \oplus A^{w_2}$ is
\begin{equation}
    w_{12}  =-\ln \left[\left(1-s_1\right)\left(1-s_2\right)\right] =-\ln \left(1-s_1\right)-\ln \left(1-s_2\right)=w_1+w_2.
\end{equation}
Hence, when aggregating simple mass functions with an identical focal set using Dempster's rule, the resulting mass function remains a \textit{simple} one, with their weights summed. This process is expressed as
\begin{equation} \label{equ:weight_add}
 A^{w_1} \oplus A^{w_2}=A^{w_1+w_2}.
\end{equation} 
Thanks to such a good property of the simple mass function, we can perform mass function fusion in two stages. The first stage involves fusing mass functions with the same focal set (e.g., $m_{11}^+(\cdot),m_{12}^+(\cdot),\dots,m_{1J}^+(\cdot)$), while the second stage fuses the mass functions obtained from the first stage. 
It is worth noting that the fusion in the first stage can be directly carried out using \eqref{equ:weight_add}, effectively avoiding substantial computational overhead.
Specifically, this can be formally expressed by
\begin{equation}
\begin{aligned}\label{equ:mj+&mj-}
    m_j^{+}(\cdot)=\bigoplus_{i=1}^I m_{ij}^{+}(\cdot)=\bigoplus_{i=1}^I \left\{z_i\right\}^{w_{ij}^{+}}=\left\{z_i\right\}^{\sum_{i}^Iw_{ij}^{+}}=\left\{z_i\right\}^{w_k^{+}};\quad j=1,2,\dots,J\\
    m_j^{-}(\cdot)=\bigoplus_{i=1}^I m_{ij}^{-}(\cdot)=\bigoplus_{i=1}^I \overline{\left\{z_i\right\}}^{w_{ij}^{-}}=\overline{\left\{z_i\right\}}^{\sum_{i}^Iw_{ij}^{-}}=\overline{\left\{z_i\right\}}^{w_k^{-}};\quad j=1,2,\dots,J\\
\end{aligned}
\end{equation}
where $w_{j}^{+}=\sum_{i=1}^{I}w_{ij}^{+}$ and $w_{j}^{-}=\sum_{i=1}^{I}w_{ij}^{-}$, and both \( m_j^+(\cdot) \) and \( m_j^-(\cdot) \) still are simple mass functions. 
It is important to note that in the first stage, the limitations inherent in DST are significantly alleviated, as shown in \eqref{equ:mj+&mj-}, by avoiding heavy calculation associated with the power set.
We can further express
\begin{equation}
    m_j^{+}(\{z_i\})=1-\exp(-w_i^+);\quad
    m_j^{-}(\overline{\{z_i\}})=1-\exp(-w_i^-).
\end{equation}
In the second stage, we apply Dempster's rule to combine these simple mass functions with different focal sets, yielding \( m^+(\cdot) \) and \( m^-(\cdot) \). 
The focal set of \( m^+(\cdot) \) is \( \mathcal V \) and \( \{z_i\} \) for \( i = 1, 2, \dots, I \). The explicit expressions of \( m^+(\cdot) \) are
\begin{equation}
\begin{aligned}
    &m^+(z_i)=\frac{\exp(w_i^-)-1}{\sum_{l=1}^I\exp(w_l^+)-I+1}\quad i=1,2\cdots,I;\\ &m^+(\mathcal V)=\frac{1}{\sum_{l=1}^I\exp(w_l^+)-I+1}
\end{aligned}
\end{equation}
It is important to note that the focal sets for \( m^-(\cdot) \) are all subsets \( A \subset \mathcal V \). In~\cite{denoeux2019logistic}, the corresponding expression can be derived as follows:
\begin{equation}
    m^{-}(A)=\frac{1}{1 - \kappa^-}\prod_{z_i\notin A}[1-\exp(-w_i^-)]\prod_{z_i\in A}\exp(-w_i^-),
\end{equation}
where \( \kappa^-=\prod_{l=1}^I[1-\exp(-w_l^-)] \) is the conflict within \( \{ m_i^-(\cdot) \}_{i=1}^I \).
Therefore, it is natural to compute the conflict between $m^+(\cdot)$ and $m^-(\cdot)$, in \cite{denoeux2019logistic}, the final expression for $\kappa$ is formalized as follows:
\begin{equation}\label{equ:kappa}
\begin{aligned}
\kappa & =\sum_{i=1}^I\left\{m^{+}\left(\left\{z_i\right\}\right) \sum_{z_i \notin A} m^{-}(A)\right\}\\&=\sum_{i=1}^I\left\{\eta^{+}\left(\exp \left(w_i^{+}\right)-1\right)\left[1-\eta^{-} \exp \left(-w_i^{-}\right)\right]\right\} .
\end{aligned}
\end{equation}
where $\eta^{+}=(\sum_{l=1}^{I}\exp(w_{l}^{+})-I+1)^{-1}$, and $\eta^{-}=(1-\prod_{l=1}^{I}[1-\exp(-w_{l}^{-})])^{-1}$ are normalization factors. 
Since LVLMs generate responses through next-token prediction, each generated token corresponds to a value of $\kappa$. To measure the feature conflict of the entire sentence, we select the maximum $\kappa$ from the tokens in the response, that is
\begin{equation}
    \kappa_{\text{max}}=\max\{\kappa^{(1)},\kappa^{(2)},\dots,\kappa^{(N)}\}
\end{equation}
where $\kappa^{(N)}$ denotes the $\kappa$ corresponding to the feature conflict associated with generating the $N$-th token. 
Furthermore, by combining \( m^+(\cdot) \) and \( m^-(\cdot) \), the overall mass function \( m(\cdot) \) is obtained, which can then be transformed into a probability mass function \( p_m(\cdot) \) through the plausibility transformation. This can be formalized as follows:
\begin{equation}\label{equ:plausibilitytransformation}
p_m(z_i)=\frac{Pl(\{z_i\})}{\sum_{l=1}^I Pl(\{z_l\})}=\frac{\sum_{B\cap\{z_i\}\not=\emptyset}m(B)}{\sum_{l=1}^I\sum_{B\cap\{z_l\}\not=\emptyset}m(B)},\quad i=1,2,\dots ,I.
\end{equation}
Essentially, this equation is equivalent to~\eqref{equ:lvlmssoftmax}, both representing the probability mass of all tokens in the vocabulary $\mathcal V$ assigned by LVLMs during a single inference round.

\section{Experiments}\label{sec:exp}

In this section, we first present the implementation details of the experiment in Subsection \ref{sec:implementdetail}. 
We then provide a comprehensive evaluation of all methods on the PRE-HAL dataset in Subsection \ref{sec:evaluation}, showing that the evidential conflict outperforms baseline uncertainty metrics in visual hallucination detection.
We show an alignment comparison of the answer correctness assessments between human evaluators and ChatGPT-4.0 in Subsection \ref{sec:gptcheck}, exploring its potential as an efficient alternative tool for multi-choice answer evaluation.

\subsection{Implementation Detail}\label{sec:implementdetail}
In this subsection, we elaborate on our experimental settings, including detailed information about the LVLMs used, baseline methods, and evaluation metrics for visual hallucination detection.
\paragraph{LVLMs under Evaluation}
We employed three state-of-the-art LVLMs, mPLUG-Owl2~\cite{ye2024mplug2}, mPLUG-Owl3~\cite{ye2024mplug3} and LLaVA-v1.5~\cite{liu2024improved}, for conducting visual hallucination detection experiments on PRE-HAL.
The mPLUG-Owl series models strengthen effectively the modality collaboration to enhance performance in both text and multimodal tasks.
Specifically, the visual encoder and language model of mPLUG-Owl2 are a Vision Transformer-Large~\cite{dosovitskiyImageWorth16x162021} (ViT-L/14) with a patch size of 14 and an LLaMA2~\cite{touvron2023llama2} with 7 billion parameters (LLaMA2-7B), respectively.
The LLaVA-v1.5~\cite{liu2024improved} is an improved version of the LVLMs LLaVA~\cite{liu2024visual}, which was tuned by multimodal language-image instruction-following data generated by GPT-4 \cite{openai2024gpt4technicalreport}.
Specifically, LLaVA-v1.5 utilizes the ViT-L/14 visual encoder from CLIP \cite{radfordLearningTransferableVisual2021} with a multi-layer projection and adopts Vicuna-v1.5-7B \cite{VicunaOpenSourceChatbot}, which is fine-tuned based on LLaMA~\cite{touvron2023llama}, as a language model.
We directly used the pretrained weights on HuggingFace\footnote{\url{https://huggingface.co/MAGAer13/mplug-owl2-llama2-7b}}\footnote{\url{https://huggingface.co/liuhaotian/llava-v1.5-7b}} without any alteration.


\paragraph{Baselines}
Since our DST-based visual hallucination detection method is an internal approach of uncertainty estimation for hallucination detection, we selected two commonly used uncertainty metrics as our baseline, predictive entropy $(PE)$~\cite{kadavath2022language} and length-normalized predictive entropy $(LN$-$PE)$~\cite{Malinin2021UncertaintyEI}.
$PE$ can be caculated by \[-\sum_{n=1}^N\sum_{z\in \mathcal V}p(z\vert z^{(<n)},x_{v},x_{t})\ln{p(z\vert z^{(<n)},x_{v},x_{t}).}\] 
$LN$-$PE$ normalizes $PE$ by sequence length $L$.
Furthermore, we also incorporated another two probability-based metrics, that is, the sum of probabilities $(PS)$ and the sum of the logarithm of probabilities $(LPS)$, as the uncertainty metrics for comparison.
Additionally, we consider the length $(L)$ of the model’s response tokens as a baseline, which was shown to be strongly correlated to the hallucinations in LLMs~\cite{nayab2024concise}.
Moreover, to enable a comprehensive comparison with existing approaches, we include two state-of-the-art external methods: Semantic Entropy (SE)\cite{kuhn2023semantic} and Self-Consistency (SC)\cite{wangself}. Both methods require repeated reasoning. SE leverages external models to assess semantic equivalence and aggregates response probabilities to compute semantic entropy. SC estimates uncertainty by analyzing the distribution of the most frequently generated outputs across multiple runs. While both SE and SC demonstrate strong performance, they incur substantial computational overhead. For fairness, we explicitly mark them in Table~\ref{tab:acc-pre-rec-f1} and Table~\ref{tab:AUROC-Dimensions}. In the subsequent analysis, we focus primarily on the proposed internal method, which does not rely on repeated generation or external models.


\paragraph{Evaluation Metrics}
 
We first used accuracy, precision, recall, and F1 score to evaluate the baselines and our method. 
When calculating these binary classification evaluation metrics, we determine the threshold of those hallucination detection metrics using the same false-positive rate for each metric.
In particular, we set the false positive rate (FPR) to 0.08. Specifically, accuracy (Acc), precision (Prec), recall (Rec), and F1 score are computed as follows:
\begin{equation}
\begin{aligned}
    &\text{Acc} = \frac{TP + TN}{TP + TN + FP + FN};\quad &\text{Prec} &= \frac{TP}{TP + FP} \\
    &\text{Rec} = \frac{TP}{TP + FN}; &\text{F1} &= 2 \cdot \frac{\text{Prec} \cdot \text{Rec}}{\text{Prec} + \text{Rec}}
\end{aligned}
\end{equation}

where \( TP \), \( TN \), \( FP \), and \( FN \) represent true positives, true negatives, false positives, and false negatives, respectively.
We use AUROC and AUPR to assess hallucination detection. AUROC is computed by varying thresholds to obtain FPR and TPR pairs, which are used to plot the ROC curve. AUPR reflects precision and recall across thresholds and is especially useful for imbalanced datasets.
To validate evidential conflict, we compute accuracy, precision, recall, and F1 score at a fixed FPR of 0.08 for each baseline method on the PRE-HAL dataset (Table 1). This setup reflects practical performance under a given threshold.
Let \(M\) samples have evidential conflicts \(\{\kappa_1, \kappa_2, \ldots, \kappa_M\}\) from Section~\ref{sec:massfusion}. For each \(\kappa_m\), we calculate FPR using
\[
\text{FPR} = \frac{\text{FP}}{\text{FP} + \text{TN}}.
\]
The threshold \(\tau_{0.08}\) is determined as the \(m^*\)-th index where
\[
m^* = \underset{m}{\arg\min} \left| \text{FPR}_m - 0.08 \right|.
\]
Samples with \(\kappa_{\text{max}} > \tau_{0.08}\) are classified as positive, and the rest as negative. Finally, we compute the metrics: accuracy, precision, recall, and F1 score.


To assess the effectiveness of these methods for visual hallucination detection, we used evaluation metrics: area under the receiver-operating-characteristic curve (AUROC) and area under the precision-recall curve (AUPR). 
To calculate the AUROC, for each of all possible thresholds, compute the corresponding False Positive Rate (FPR) and True Positive Rate (TPR). Subsequently, the Receiver Operating Characteristic (ROC) curve is plotted using all combinations of FPR and TPR, and the AUROC.
The AUPR represents the Precision and Recall values at different  thresholds. This metric offers a more comprehensive and accurate means of comparing detection methods, especially when dealing with imbalanced datasets.
\paragraph{Hardware} All experiments were performed on NVIDIA L20 GPUs.
\begin{table}[t!]
    \centering
    \footnotesize
    \caption{Detection performance of three LVLMs on the PRE-HAL dataset. The best results in each group are indicated in \textbf{bold}, and the results of $\kappa_{\text{max}}$ that exceed the external methods in each group are marked in \textcolor{blue}{blue}.}
    \begin{tabular}{lc|c|ccc|c}
        \toprule
        &LVLM&Metric&Accuracy&Precision&Recall&F1 Score\\
        \midrule
        & \multirow{10}{*}{LLaVA-v1.5} &\multicolumn{5}{c}{External Method (Baselines)}\\
        \cline{3-7}
        &&$SE$ & $74.96$ & $45.10$ & \textbf{15.47} & \textbf{23.04}  \\
        &&$SC$ & \textbf{75.14} & \textbf{46.30} & $9.42$ & $15.65$  \\
        \cline{3-7}
        &&\multicolumn{5}{c}{Internal Method}\\
        \cline{3-7}
        && $PS$ & $74.40$ & $40.30$ & $17.03$ & $23.95$ \\
       & & $LPS$ & $66.84$ & $1.33$ & $0.55$ & $0.78$\\
        &&$PE$ & $74.36$ & $40.10$ & $16.95$ & $23.83$  \\
        &&$LN$-$PE$ & $66.83$ & $1.23$ & $0.51$ & $0.72$  \\
        &&$L$ & $74.15$ & $40.00$ & \textbf{18.51} & $25.31$ \\
        
        &&$\kappa_{\text{max}}$ & \textbf{74.94} & \textbf{43.00} & \textcolor{blue}{18.17} & \textcolor{blue}{\textbf{25.55}} \\
        \midrule
       & \multirow{10}{*}{mPLUG-Owl2}&\multicolumn{5}{c}{External Method (Baselines)}\\
        \cline{3-7}
       &&$SE$ & $75.42$ & $41.21$ & \textbf{17.94} & \textbf{25.00}  \\
        &&$SC$ & \textbf{76.17} & \textbf{48.30} & $10.45$ & $17.18$ \\
        \cline{3-7}
        &&\multicolumn{5}{c}{Internal Method}\\
        \cline{3-7}
       && $PS$ & $75.01$ & $38.60$ & $17.00$ & $23.60$  \\
        &&$LPS$ & $69.77$ & $12.32$ & $5.42$ & $7.53$  \\
        &&$PE$ & $75.10$ & $39.04$ & $17.17$ & $23.85$ \\
        &&$LN$-$PE$ & $72.68$ & $14.26$ & $4.05$ & $6.31$\\
        &&$L$ &$75.36$ & $40.08$ & $17.17$ & $24.04$ \\
         
        &&$\kappa_{\text{max}}$ & \textcolor{blue}{\textbf{76.67}} & \textbf{46.90} & \textcolor{blue}{\textbf{20.65}} & \textcolor{blue}{\textbf{28.68}}  \\
        \midrule
       & \multirow{10}{*}{mPLUG-Owl3}&\multicolumn{5}{c}{External Method (Baselines)}\\\cline{3-7}
       &&$SE$ & $81.75$ & $41.00$ & \textbf{24.92} & \textbf{31.00}  \\
        &&$SC$ & \textbf{83.90} & \textbf{56.60} & $9.12$ & $15.71$  \\
        \cline{3-7}
        &&\multicolumn{5}{c}{Internal Method}\\
        \cline{3-7}
       && $PS$ & $77.05$ & $17.50$ & $10.64$ & $13.23$  \\
        &&$LPS$ & $73.85$ & $1.50$ & $0.91$ & $1.13$  \\
        &&$PE$ & $77.15$ & $17.68$ & $10.64$ & $13.28$ \\
        &&$LN$-$PE$ & $73.95$ & $2.00$ & $1.22$ & $1.51$\\
        &&$L$ &$75.10$ & $25.18$ & $10.64$ & $14.96$ \\
         
        &&$\kappa_{\text{max}}$ &\textbf{78.75} & \textbf{26.00} & \textbf{15.81} & \textbf{19.66}  \\
        
        \bottomrule
    \end{tabular}
\label{tab:acc-pre-rec-f1}
\end{table}

\subsection{Visual Hallucination Detection on PRE-HAL}\label{sec:evaluation}
To validate the effectiveness of evidential conflict, we first present the accuracy, precision, recall, and F1 score for each baseline method at a fixed false-positive rate of 0.08 on the visual hallucination detection task in the PRE-HAL dataset, as shown in Table~\ref{tab:acc-pre-rec-f1}. 
This experimental setup holds great practical significance, as it directly reflects the performance of hallucination detection under a given threshold.
Our method achieves optimal performance across all metrics on mPLUG-Owl2 and mPLUG-Owl3. In contrast, on LLaVA-v1.5, it outperforms all methods in every metric except the recall rate.
Moreover, we observe that the length $(L)$ of the model’s response tokens, which is straightforward and easy to compute, also yields satisfactory detection results. This finding is consistent with the conclusion in \cite{nayab2024concise}, and intuitively, the probability of hallucination increases as the amount of model output grows.
An interesting observation is that, although LPS and LN-PE achieve satisfactory results in terms of accuracy, their performance on precision, recall, and F1 score is suboptimal. This may be attributed to the length normalization applied by both methods. This suggests that when designing uncertainty-driven visual hallucination detection methods, the impact of length should be taken into consideration.
Additionally, we observe that both the baselines and the proposed method tend to achieve higher scores on mPLUG-Owl2 compared to their corresponding values on LLaVA, which can be attributed to the better calibration of mPLUG-Owl2.

Since Table~\ref{tab:acc-pre-rec-f1} only presents results with a fixed false-positive rate, we conducted more comprehensive experiments to further validate the effectiveness of our method with enhanced statistical significance. 
These results, shown in Table~\ref{tab:AUROC-Dimensions}, present the AUROC scores for visual hallucination detection on the PRE-HAL dataset, comparing our method with five baseline approaches.
As shown in the last column of Table~\ref{tab:AUROC-Dimensions}, our method outperforms the best baseline on the three LVLMs, respectively, particularly with a 10\% improvement on the well-calibrated mPLUG-Owl2.
The length $(L)$ of the model’s response tokens yields promising results, consistent with the observations in Table~\ref{tab:acc-pre-rec-f1}. The performance of all methods is superior on mPLUG-Owl2. Consequently, we speculate that these metrics tend to perform better on models that exhibit stronger overall performance.
It is noteworthy that while some external methods, such as Self-Consistency (SC) and Semantic Entropy (SE), demonstrate strengths in specific metrics or hallucination types, our proposed method achieves competitive results in many cases, as highlighted in blue in Table~\ref{tab:acc-pre-rec-f1} and Table~\ref{tab:AUROC-Dimensions}. Notably, our approach eliminates the need for multiple inferences or external model calls, yet still delivers strong performance, emphasizing its efficiency and practical applicability in real-world scenarios.

To further analyze, we conducted a more detailed comparison of these methods across different types of hallucinations. It has been shown that evidential conflict outperforms other methods across all types of visual hallucinations.

\begin{figure}[ht!]
    \centering
   
     \subfloat[LLaVA-v1.5 AUROC]{%
        \begin{minipage}[b]{0.32\textwidth}
            \centering
            \includegraphics[width=\textwidth,clip, trim={0cm 0cm 0cm 0cm}]{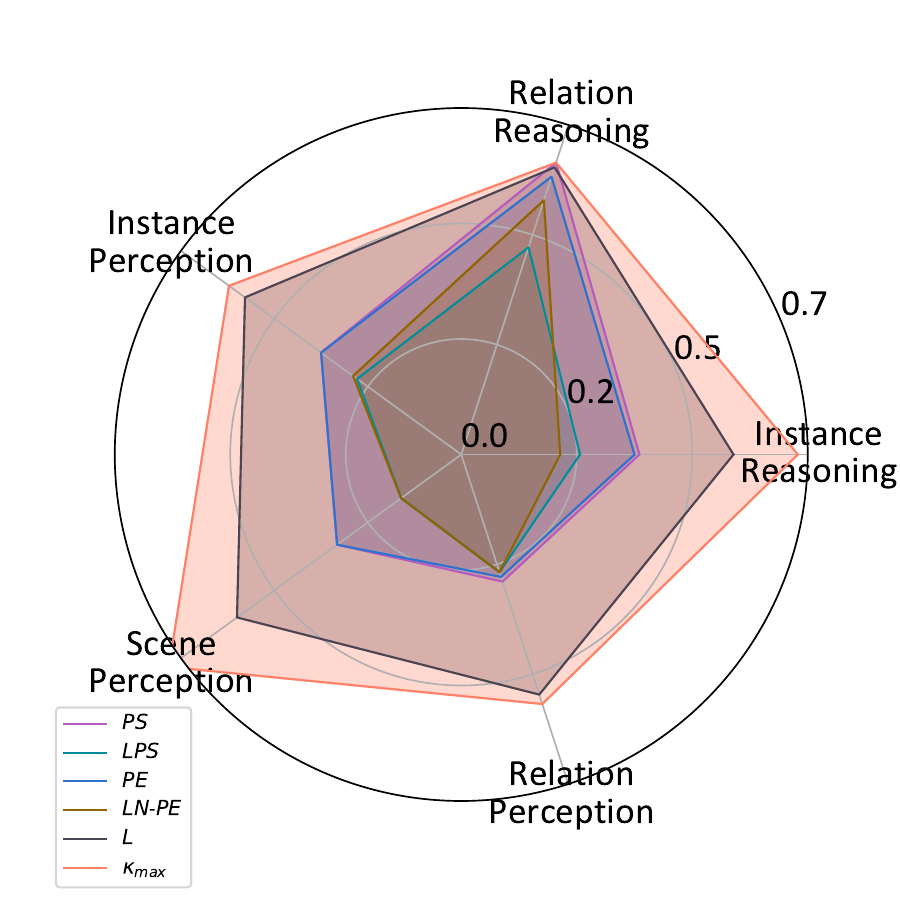}
        \end{minipage}%
        \label{fig:Radar-LLaVA-auroc}%
    }
    \hspace{0cm} 
    \subfloat[mPLUG-Owl2 AUROC]{%
        \begin{minipage}[b]{0.32\textwidth}
            \centering
            \includegraphics[width=\textwidth,clip, trim={0cm 0cm 0cm 0cm}]{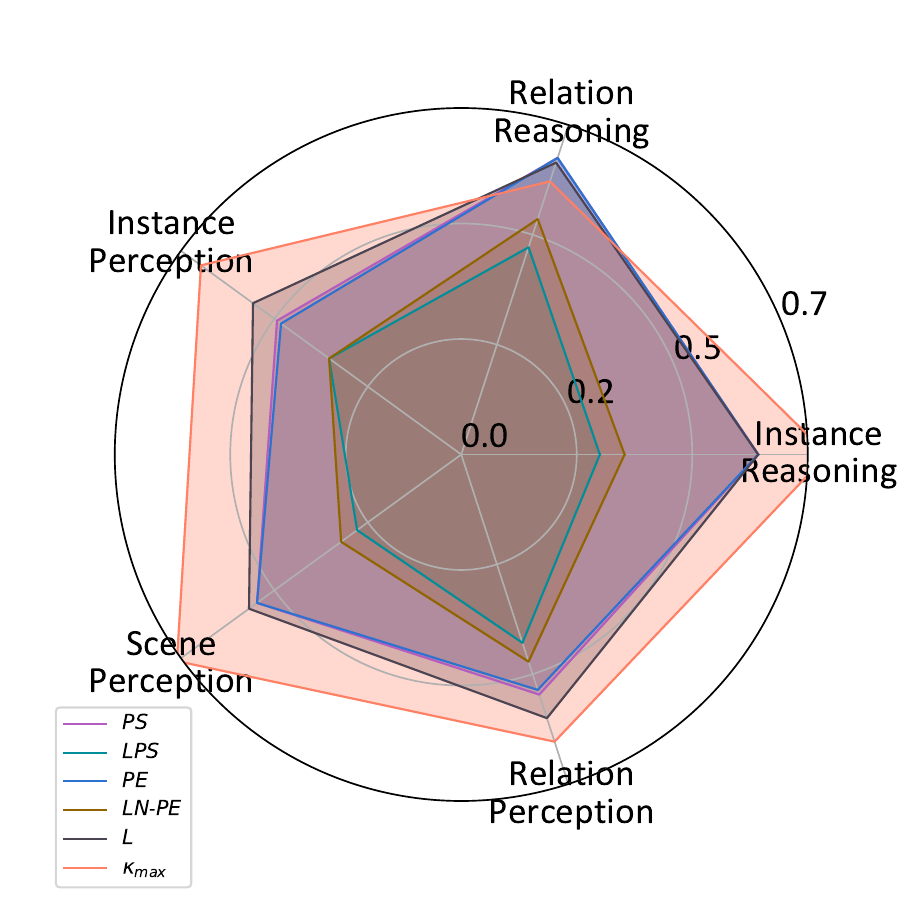}
        \end{minipage}%
        \label{fig:Radar-Owl2-auroc}%
    }
    \hspace{0cm} 
    \subfloat[mPLUG-Owl3 AUROC]{%
        \begin{minipage}[b]{0.32\textwidth}
            \centering
            \includegraphics[width=\textwidth,clip, trim={0cm 0cm 0cm 0cm}]{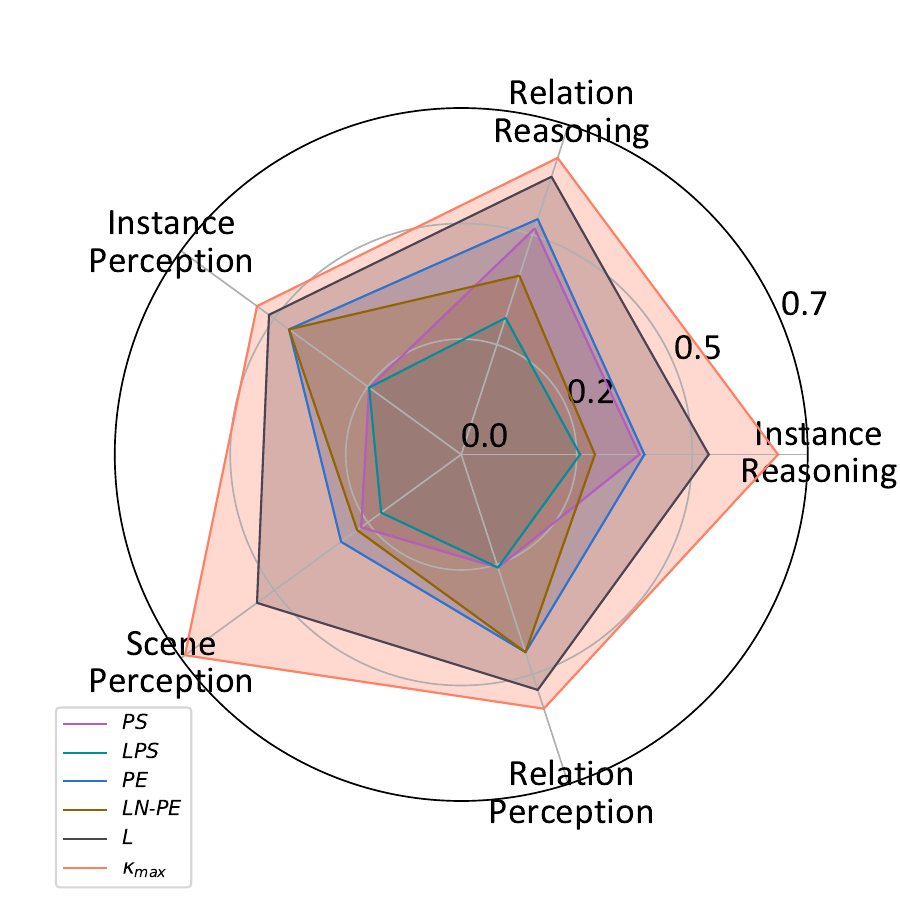}
        \end{minipage}%
        \label{fig:Radar-Owl3-auroc}%
    }

    \vspace{0cm} 
    
        \subfloat[LLaVA-v1.5 AUPR]{%
        \begin{minipage}[b]{0.32\textwidth}
            \centering
            \includegraphics[width=\textwidth,clip, trim={0cm 0cm 0cm 0cm}]{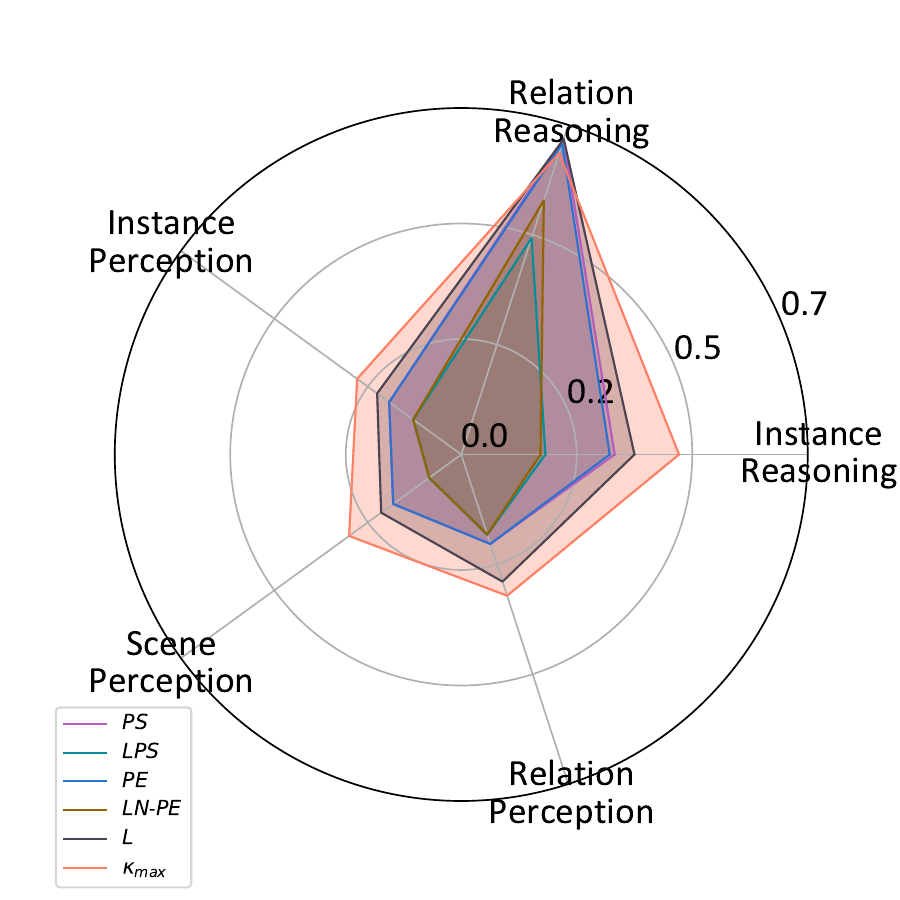}
        \end{minipage}%
        \label{fig:Radar-LLaVA-aupr}%
    }
    \hspace{0cm} 
    \subfloat[mPLUG-Owl2 AUPR]{%
        \begin{minipage}[b]{0.32\textwidth}
            \centering
            \includegraphics[width=\textwidth,clip, trim={0cm 0cm 0cm 0cm}]{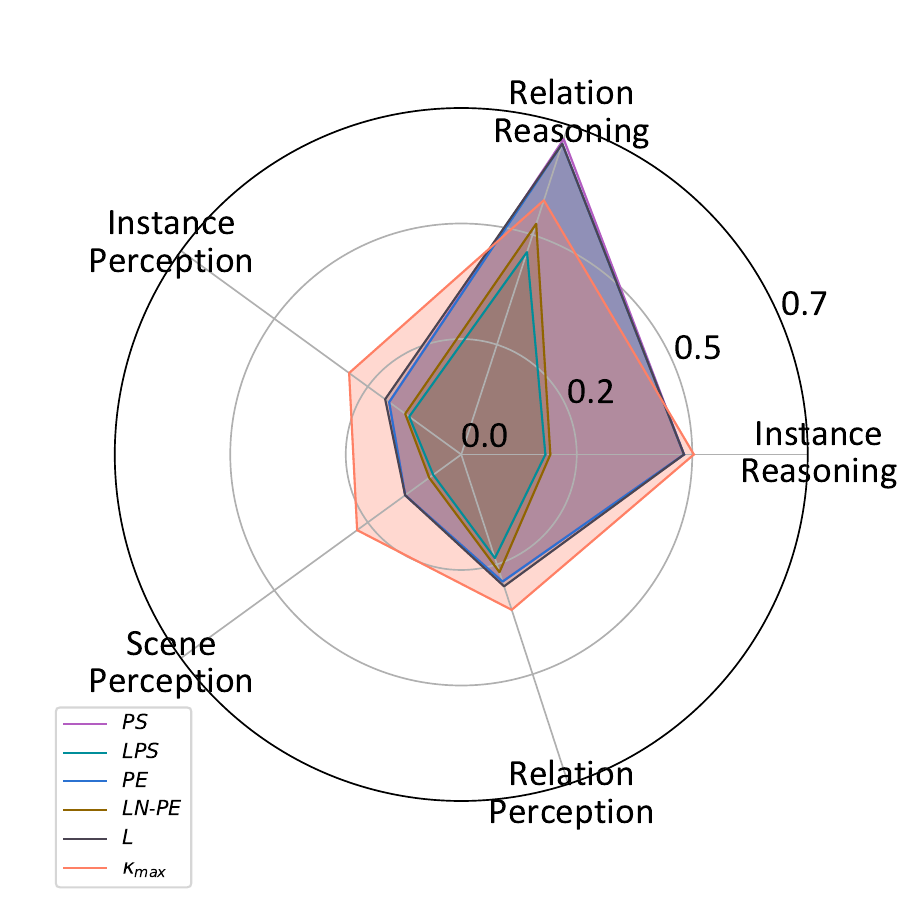}
        \end{minipage}%
        \label{fig:Radar-Owl2-aupr}%
    }
    \hspace{0cm} 
    \subfloat[mPLUG-Owl3 AUPR]{%
        \begin{minipage}[b]{0.32\textwidth}
            \centering
            \includegraphics[width=\textwidth,clip, trim={0cm 0cm 0cm 0cm}]{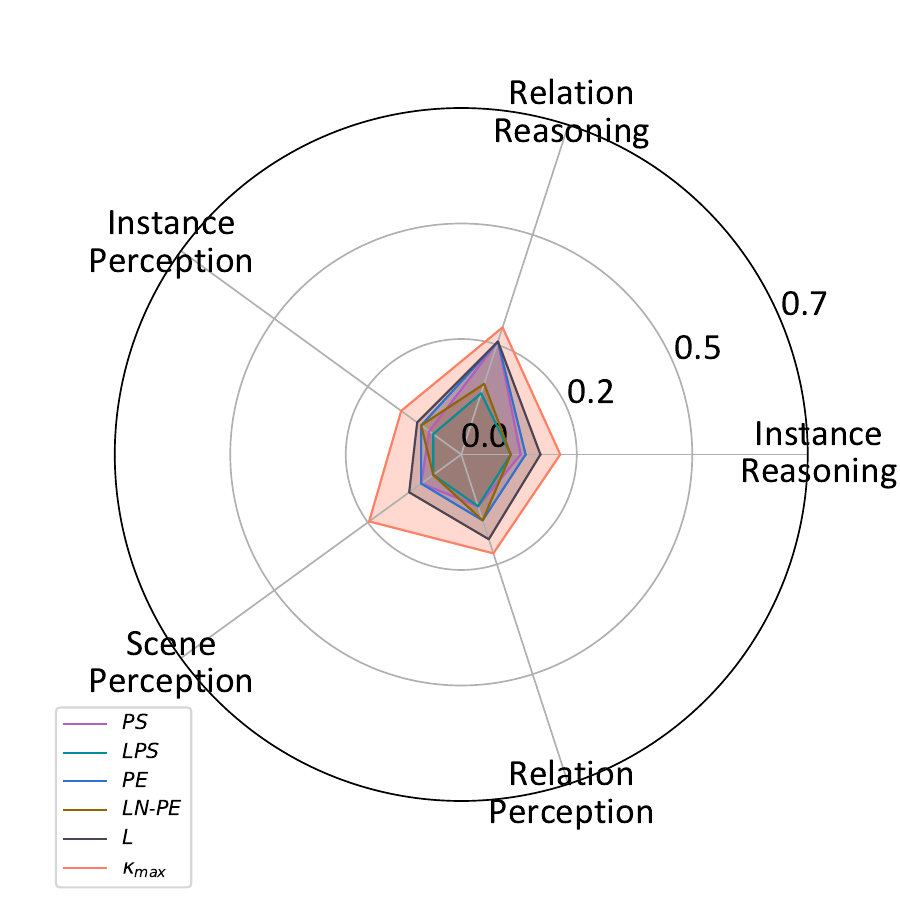}
        \end{minipage}%
        \label{fig:Radar-Owl3-aupr}%
    }
    \caption{Visual hallucination detection rates across the five hallucination types on the three LVLMs. }
    \label{fig:subfloat-combined-four}
\label{fig:Radar}
\end{figure}

Figure~\ref{fig:Radar} presents the overall detection rates for the five types of visual hallucinations on the three LVLMs, based on both AUROC and AUPR.

\begin{table}[t!]
    \centering
    \footnotesize
    \setlength\tabcolsep{4pt}
    \renewcommand{\arraystretch}{0.9}
    \caption{AUROC scores on all categories of visual hallucinations for three LVLMs. The best results in each group are indicated in \textbf{bold}, and the results of $\kappa_{\text{max}}$ that exceed the external methods in each group are marked in \textcolor{blue}{blue}.}
    \begin{tabular}{lc|c|cc|ccc|c}
        \toprule
        &\multirow{2}{*}{LVLM}&\multirow{2}{*}{Metric}&\multicolumn{2}{c|}{Model Capability}&\multicolumn{3}{c|}{Visual Semantics}&\multirow{2}{*}{Total}\\
        \cline{4-8}
       && & Perception & Reasoning & Instance &Scene & Relation &\rule{0pt}{3ex} \\
       \midrule
        & \multirow{10}{*}{LLaVA-v1.5} &\multicolumn{7}{c}{External Method (Baselines)}\\
        \cline{3-9}
        &&$SE$ & $0.59$ & $0.56$ & $0.59$& $0.72$& $0.60$ &$0.61$ \\
        &&$SC$ &\textbf{0.65} & \textbf{0.57} & \textbf{0.65}& \textbf{0.75}& \textbf{0.64} &\textbf{0.65} \\
        \cline{3-9}
        &&\multicolumn{7}{c}{Internal Method}\\
        \cline{3-9}
        && $PS$ & $0.30$ & $0.58$ & $0.35$&$0.31$ & $0.40$&$0.36$ \\
       & & $LPS$ & $0.25$ & $0.34$ & $0.26$ &$0.15$& $0.26$&$0.24$ \\
        &&$PE$ & $0.31$ & $0.55$ & $0.35$&$0.31$ & $0.38$ &$0.36$ \\
        &&$LN$-$PE$ & $0.25$ & $0.41$ & $0.26$&$0.15$ & $0.30$ & $0.26$ \\
        &&$L$ & $0.52$ & $0.63$ & $0.54$ &$0.56$& $0.57$ &$0.55$\\
        
        &&$\kappa_{\text{max}}$ & \textbf{0.55} & \textcolor{blue}{\textbf{0.68}} & \textbf{0.59}& \textbf{0.71}& \textbf{0.60} &\textbf{0.59}\\
        \midrule
       & \multirow{10}{*}{mPLUG-Owl2}&\multicolumn{7}{c}{External Method (Baselines)}\\
       \cline{3-9}
       &&$SE$ & $0.61$ & $0.60$ & $0.63$& $0.60$& $0.65$ &$0.63$ \\
        &&$SC$ & \textbf{0.63} & \textbf{0.61} & \textbf{0.66}& \textbf{0.61}& \textbf{0.66} &\textbf{0.66} \\
        \cline{3-9}
        &&\multicolumn{7}{c}{Internal Method}\\
        \cline{3-9}
       && $PS$ & $0.47$ & $0.65$ & $0.48$&$0.51$ & $0.60$ &$0.54$ \\
        &&$LPS$ & $0.35$ & $0.33$ & $0.33$& $0.26$& $0.37$ &$0.34$ \\
        &&$PE$ & $0.46$ & $0.65$ & $0.48$ &$0.51$& $0.60$ &$0.54$ \\
        &&$LN$-$PE$ & $0.36$ & $0.40$ & $0.36$&$0.30$ & $0.55$ &$0.44$ \\
        &&$L$ & $0.52$& $0.65$ & $0.53$&$0.53$ & $0.63$ &$0.57$ \\
        
        &&$\kappa_{\text{max}}$ & \textcolor{blue}{\textbf{0.64}} & \textcolor{blue}{\textbf{0.67}} & \textcolor{blue}{\textbf{0.66}} &\textcolor{blue}{\textbf{0.71}}& \textbf{0.65} &\textcolor{blue}{\textbf{0.67}} \\
        
        \midrule
       & \multirow{10}{*}{mPLUG-Owl3}&\multicolumn{7}{c}{External Method (Baselines)}\\
       \cline{3-9}
       &&$SE$ & $0.59$ & \textbf{0.75} & $0.58$& $0.73$& $0.66$ &$0.64$ \\
        &&$SC$ & \textbf{0.63} & \textbf{0.75} & \textbf{0.64}& \textbf{0.79}& \textbf{0.68} &\textbf{0.68} \\
        \cline{3-9}
        &&\multicolumn{7}{c}{Internal Method}\\
        \cline{3-9}
       && $PS$ & $0.25$ & $0.44$ & $0.25$&$0.25$ & $0.32$ &$0.29$ \\
        &&$LPS$ & $0.24$ & $0.27$ & $0.24$& $0.20$& $0.27$ &$0.25$ \\
        &&$PE$ & $0.42$ & $0.45$ & $0.44$ &$0.30$& $0.47$ &$0.44$ \\
        &&$LN$-$PE$ & $0.42$ & $0.33$ & $0.42$&$0.26$ & $0.42$ &$0.41$ \\
        &&$L$ & $0.50$& $0.56$ & $0.51$&$0.51$ & $0.52$ &$0.51$ \\
        
        &&$\kappa_{\text{max}}$ &\textbf{0.53} & \textbf{0.64} & \textbf{0.54} &\textbf{0.69}& \textbf{0.56} &\textbf{0.58} \\
        \bottomrule
    \end{tabular}
    
\label{tab:AUROC-Dimensions}
\end{table}

\begin{figure}[htbp]
    \hspace{0cm} 
    \centering
    \subfloat{%
        \begin{minipage}[b]{0.49\textwidth}
            \centering
            
                \includegraphics[width=\textwidth,clip, trim={0cm 0cm 0cm 0cm}]{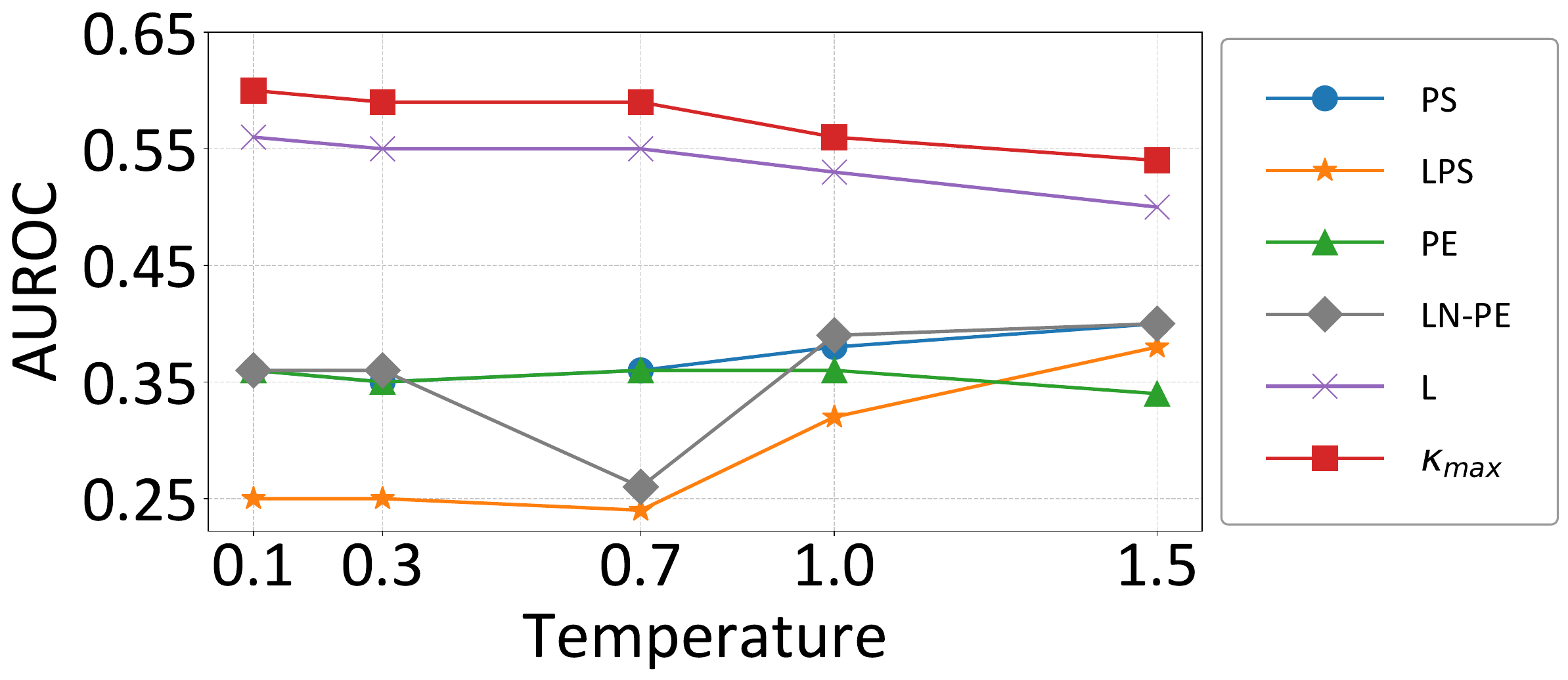}%
            
        \end{minipage}%
        \label{fig:temperature}%
    }
    \hspace{0cm} 
    \subfloat{%
        \begin{minipage}[b]{0.49\textwidth}
            \centering
            \includegraphics[width=\textwidth,clip, trim={0cm 0cm 0cm 0cm}]{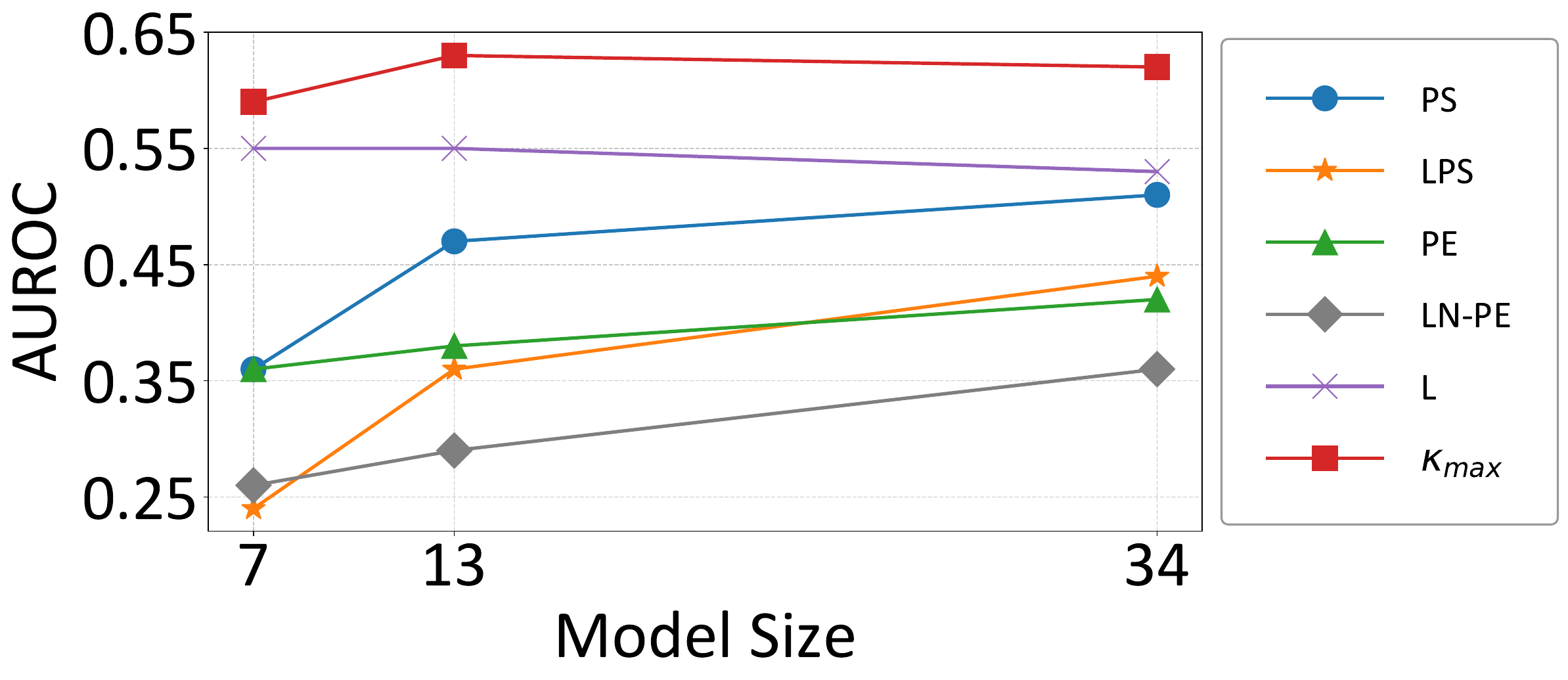}
        \end{minipage}%
        \label{fig:modelsize}%
    }
    \caption{Ablation study of AUROC on LLaVA-v1.5 under varying temperatures and model sizes.}
    \label{fig:ablation}
\end{figure}
\subsection{Ablation Study}\label{sec:ablation}
In this section, we conduct ablation studies to evaluate the effectiveness of our method by analyzing how temperature and model size influence the performance of various metrics.

\paragraph{Temperature}
We analyze the effect of temperature during generation in LVLMs, as shown in the left panel of Figure~\ref{fig:ablation}. A higher temperature generally produces more diverse outputs. We evaluate five temperature settings: $0.1$, $0.3$, $0.7$, $1.0$, and $1.5$, using LLaVA-v1.5-7B. While the AUROC changes with temperature, the $\kappa_{\text{max}}$ consistently performs best across all settings. Notably, the performance of the evidence conflict metric remains stable across different inference temperatures.

\paragraph{Model Size}
The impact of model size is illustrated in the right panel of Figure~\ref{fig:ablation}. We compare models with 7B, 13B, and 34B parameters. As LLaVA-v1.5 is only available in 7B and 13B configurations, we adopt LLaVA-v1.6 for the 34B setting to complete the comparison. Across all model sizes, $\kappa_{\text{max}}$ consistently outperforms the baselines. Compared to other internal methods, our uncertainty quantification metric remains more stable across the three model sizes (except for length $L$), with a slight performance improvement observed at the 13B parameter setting.

\subsection{Investigation of LLM-Based Result Check}\label{sec:gptcheck}
\begin{figure}
\centering
\includegraphics[width=0.8\linewidth,clip, trim={0cm 1cm 0cm 0cm}]{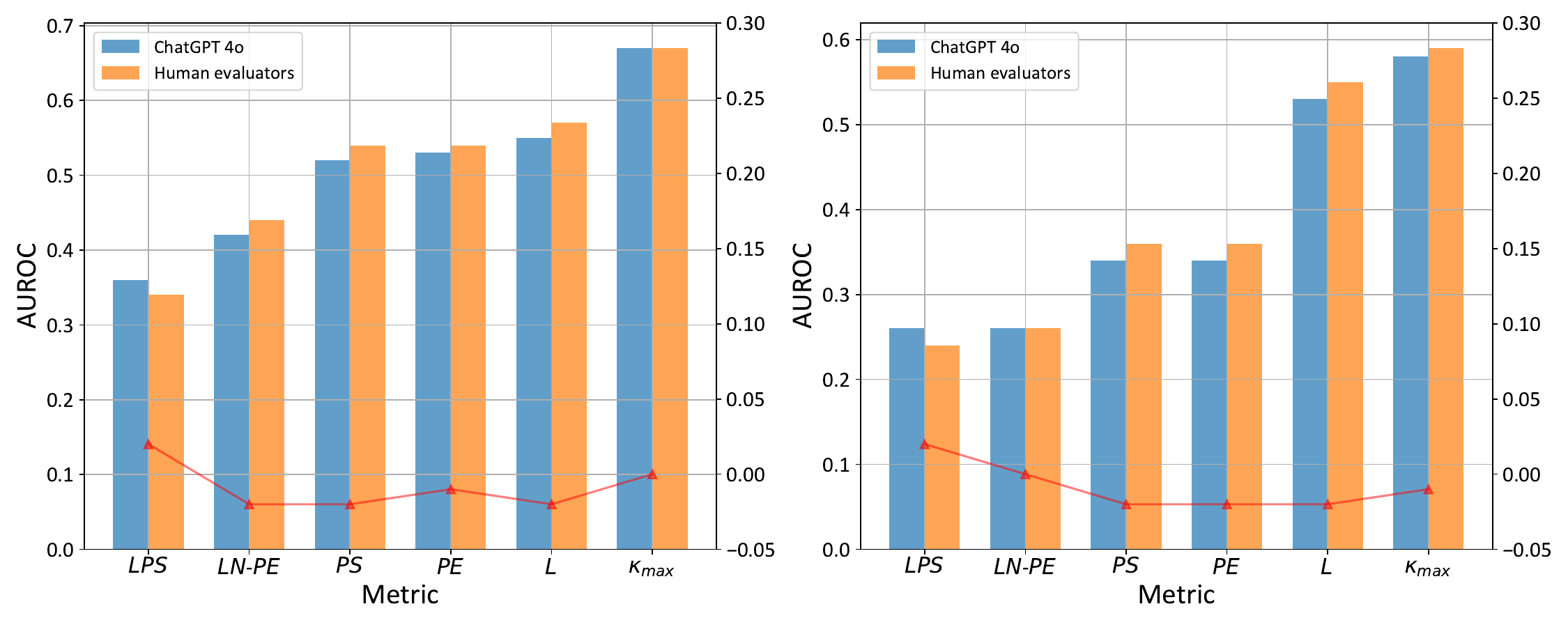}
\caption{AUROC scores based on the correctness results of LVLM outputs assessed by ChatGPT 4o and human evaluators. Results correspond to the tests conducted on mPLUG-Owl2 and LLaVA-v1.5, respectively.}\label{fig:GPT}
\end{figure}
Assessing the correctness of model outputs with ground-truth answers by humans is highly resource-intensive and unsustainable. Thus, we explore automated methods for this assessment, which will improve overall process efficiency and provide actionable insights for the community. 

We called the API of ChatGPT 4o to assess the correctness of the inference results of the LVLMs. To ensure the reliability of the evaluation results for ChatGPT 4o, we verified these judgments by human volunteers.
Then, we compared the results assessed by ChatGPT 4o with those by volunteers.
As shown in Figure~\ref{fig:GPT}, the differences between them were not large enough to impact the experimental results.
Therefore, we empirically show that, compared to the resource-intensive human assessment, ChatGPT 4o offers an efficient alternative for multiple-choice question evaluation with an acceptable deviation range.

\section{Discussion}
\paragraph{The Applicability of Our Method}
Although this paper focuses on detecting visual hallucinations in LVLMs for VQA tasks, our method calculates evidential uncertainty using intermediate features and the parameters of the FFN, and it can also be adapted to detect other types of prediction errors in various models that incorporate FFNs. For example, it has been validated for misprediction detection in classification networks~\cite{denoeux2019logistic}, as well as in DeepLabV3-Plus for semantic segmentation~\cite{wang2025reliable}.
In addition, uncertainty quantification methods are effective for identifying abnormal examples, such as adversarial inputs and OOD data~\cite{sensoy2018evidential,aguilar2023continual,franchi2024make,beechey2023evidential}, and our approach can be extended to these scenarios as well.
It is important to note that our method is applicable only to open-source models, since it requires access to model parameters and intermediate features. Therefore, it cannot be applied to black-box models that are accessible only through APIs. On the other hand, as a plug-and-play approach, our method does not require any additional model training, making it more efficient in terms of computational cost.

\section{Conclusion}

We propose a novel visual hallucination detection method for LVLMs through evidential conflict estimation in high-level features. Our approach innovatively leverages the transformer decoder's final layer outputs as evidence sources to quantify token-level uncertainties and aggregate them for visual hallucination detection. In the meantime, we introduced PRE-HAL benchmark, which enables systematic evaluation of both perceptual and reasoning capabilities across five hallucination types, revealing critical vulnerabilities in relation to reasoning tasks where state-of-the-art LVLMs exhibit hallucination rates exceeding 49\%.

Extensive experiments highlight the superiority of our method over conventional uncertainty metrics, achieving average AUROC improvements of 4\% to 10\% across different models. 
Notably, the method shows exceptional robustness in scene perception tasks (AUROC: 71-73\%), significantly outperforming entropy-based and probability-based baselines. This implies that this DST-based approaches effectively circumvent the calibration limitations of probability-based methods. Our work establishes feature-level conflict analysis as a computationally efficient, inference-phase solution for hallucination detection, requiring no architectural modifications or multiple forward passes.



\bibliographystyle{plain} 
\bibliography{IJAR}











\end{document}